\documentclass[nohyperref]{article}
\usepackage{hyperref}
\usepackage[accepted]{icml2022}
\newif\ifready
\readyfalse
\newcommand{\E}{\mathbb{E}}

\newcommand{\KL}{D_{\mathrm{KL}}}

\newcommand{\CE}{\mathrm{CE}}

\newcommand{\pit}{{\pi_\theta}}

\newcommand{\EX}[1]{\E_{#1}}

\newif\ifready

\readyfalse

\ifready
\newcommand{\ftk}[1]{}
\newcommand{\fgk}[1]{}
\newcommand{\fmd}[1]{}
\newcommand{\fhe}[1]{}
\newcommand{\annot}[1]{}
\newcommand{\todo}[1]{}

\else
\newcommand{\ftk}[1]{}
\newcommand{\fgk}[1]{}
\newcommand{\fmd}[1]{}
\newcommand{\fhe}[1]{}

\newcommand{\annot}[1]{\textbf{\textcolor{purple}{#1~}}}
\newcommand{\todo}[1]{\textcolor{red}{#1~}}

\fi

\newcommand{\ptmd}[1]{{#1}}

\newcommand{\NER}{\mathrm{NER}}
\usepackage[utf8]{inputenc} 
\usepackage[T1]{fontenc}    
\usepackage{hyperref}       
\usepackage{url}            
\usepackage{booktabs}       
\usepackage{amsmath,amsfonts,bm}
\usepackage{multirow}

\usepackage{tikz}
\usepackage{array}
\usepackage{xcolor}
\usepackage{calc}
\def\basiceval#1{\the\numexpr#1\relax}
\usepackage{float}

\usepackage[font=small,labelfont=bf]{caption}
\usepackage{tabularx} 
\usepackage{setspace}
\usepackage{enumitem}
\setitemize{noitemsep,topsep=0pt,parsep=0pt,partopsep=0pt}
\usepackage{xspace}
\usepackage{comment}
\usepackage{enumitem}
\usepackage[noend]{algpseudocode}
\usepackage{textcomp}
\usepackage{wrapfig}
\usepackage{enumitem}
\usepackage{bbm}
\usepackage{subcaption}
\usepackage{caption}
\usepackage[frozencache,cachedir=.]{minted}
\renewcommand{\algorithmicrequire}{\textbf{Input:}}
\renewcommand{\algorithmicensure}{\textbf{Output:}}

\begin{document}

\twocolumn[
\icmltitle{Controlling Conditional Language Models without Catastrophic Forgetting}

\icmlsetsymbol{equal}{*}

\begin{icmlauthorlist}
\icmlauthor{Tomasz Korbak}{equal,sussex}
\icmlauthor{Hady Elsahar}{naver}
\icmlauthor{Germán Kruszewski}{naver}
\icmlauthor{Marc Dymetman}{naver}

\end{icmlauthorlist}

\icmlaffiliation{sussex}{University of Sussex}
\icmlaffiliation{naver}{Naver Labs Europe}
\icmlcorrespondingauthor{Tomasz Korbak}{tomasz.korbak@gmail.com}

\icmlkeywords{controllable language generation, energy-based models, catastrophic forgetting, summarization, code generation}

\vskip 0.3in
]

\printAffiliationsAndNotice{*Work done during an internship at Naver Labs Europe.} 
\begin{abstract}
  Machine learning is shifting towards general-purpose pretrained generative models, trained in a self-supervised manner on large amounts of data, which can then be applied to solve a large number of tasks.
  However, due to their generic training methodology, these models often fail to meet some of the downstream requirements (e.g., hallucinations in abstractive summarization or style violations in code generation). This raises the important question of how to adapt pre-trained generative models to meet all requirements without destroying their general capabilities (``catastrophic forgetting''). 
  Recent work has proposed to solve this problem by representing task-specific requirements through energy-based models (EBMs) and approximating these EBMs using distributional policy gradients (DPG). Despite its effectiveness, this approach is however limited to unconditional distributions. 
  In this paper, we extend DPG to conditional tasks by proposing Conditional DPG (CDPG). We evaluate CDPG on four different control objectives across three tasks (translation, summarization and code generation) and two pretrained models (T5 and GPT-Neo). Our results show that fine-tuning using CDPG robustly moves these pretrained models closer towards meeting control objectives and --- in contrast with baseline approaches --- does not result in catastrophic forgetting.
\end{abstract}

\section{Introduction}

\begin{figure}[t]
\begin{center}
\centerline{\includegraphics[width=0.98\columnwidth]{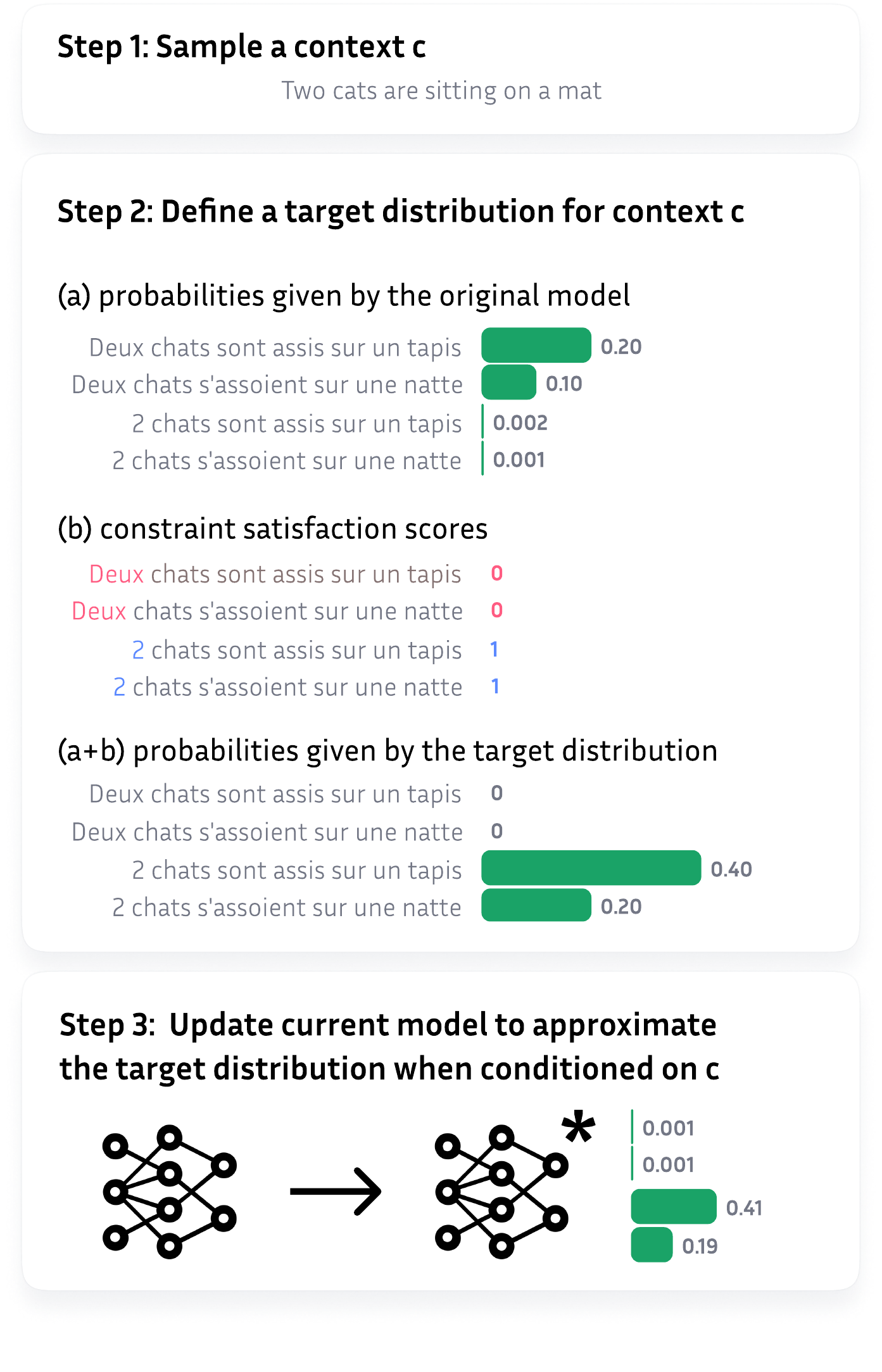}}
\vspace{-10px}
\caption{\small{An overview of our algorithm for fine-tuning conditional language models, illustrated on the terminology-constrained translation task (see Section \ref{experiments:translation} for details).}}
\end{center}
\vspace{-30px}
\end{figure}

Pretrained generative models are shifting the landscape of machine learning research and practice. General purpose models such as the GPT family ~\citep{radford2019language,gpt3,gpt-neo}, T5~\citep{2020t5}, CLIP~\citep{radford2021learning} and Codex~\citep{chen2021evaluating} are trained in a self-supervised manner on large amounts of uncurated data and can then be adapted to specific downstream tasks (e.g. generating Python code) or control objectives (e.g. controlling the style of generated code). Frequently, control objectives are motivated by the desire to address the shortcomings of certain pretrained models. These can be due to the uncurated nature of the original training data (e.g. a large portion of Python source code on the Internet violates PEP8 \citep{pep8}, the Python Style Guide) or the difficulty of learning a desired behaviour by purely self-supervised training (e.g. there is not enough training signal to ensure that a model trained on source code always generates compilable code or that a summarization model always produces factually correct summaries).

The practice of adapting and controlling pretrained generative models poses two open problems. The first is that control objectives frequently lack ground truth data that could be used for supervised fine-tuning; in general, we are only given an indicator $b(x)$ of whether a given sample $x$ from the model satisfies a given control objective. Thus, this restriction calls for approaches that can employ this signal such as reinforcement learning (RL) \citep{pasunuru-bansal-2018-multi,Ziegler19}, weighted decoding \citep{ghazvininejad-etal-2017-hafez,learning2write-holtzman-2018,see-etal-2019-makes} or decoding with perturbed activations \citep{plug_and_play_20}. 

The second problem is \emph{catastrophic forgetting}: most approaches to enforcing a control objective result in a dramatic loss of capabilities of the original model beyond the scope of the control objective. Notably, there exists one approach largely avoiding both of these problems \citep{A-parshakova-etal-2019-global,khalifa_2021}:
representing the control objective as an energy-based models (EBM) and approximating that EBM using distributional policy gradients (DPG). 
Although this approach shows great improvements in controlling pretrained language models while avoiding catastrophic forgetting~\citep{khalifa_2021}, it is limited to unconditional generation tasks and cannot fine-tune conditional models that are behind the most impactful NLP tasks such as machine translation, summarization or dialogue systems.

In this paper, we present Conditional DPG (CDPG), an extension of DPG that can approximate \emph{conditional} EBMs. A conditional EBM $\mathcal{P}$ defines an unnormalized distribution for each context $c$ (e.g. a source document). Extending the approach of \citet{khalifa_2021} \ptmd{such a conditional EBM  represents} the ideal behaviour of a generative model, given context $c$, as the distribution that incorporates the control objectives while remaining as close as possible to the original distribution to prevent catastrophic forgetting. 
%
This corresponds to defining multiple distributions $p_c$ indexed by 
$c$, \ptmd{where} each 
\ptmd{is the normalization of an unconditional}
EBM $P_c$ following \citet{khalifa_2021}. We then define the training objective for the conditional model based on minimizing the \emph{average} divergence \ptmd{for} each 
\ptmd{$p_c$}.

We demonstrate the effectiveness of CDPG in addressing shortcomings of pretrained generative models by considering three tasks: translation, summarization and code generation; and two corresponding pretrained generative models: T5 \citep{2020t5} and GPT-Neo \citep{gpt-neo}. 

We start by demonstrating the effectiveness of CDPG on a toy control objective for translation: ensuring that numeral nouns (e.g. ``two'') are translated as digits (e.g. ``1'') while other aspects of translation are unchanged. This problem is an simple instance of broader challenge of incorporating prior information in neural translation models. CDPG is able to make samples satisfying the constraint 116 times more likely. 

For summarization, a similar big, unsolved problem is ensuring that summaries are factually faithful to source documents given that summarization models are prone to hallucinating named entities never mentioned in the source \citep{maynez-etal-2020-faithfulness}. We show that a preference for factually faithful summaries (operationalized as entity-level factual consistency \citep{nan-etal-2021-entity}) can be represented by a conditional EBM. Then, we show that using CDPG to fine-tune T5 to approximate this EBM increases the number of correct and relevant named entities in summaries and improves T5's Rouge score. In contrast with RL approaches, CDPG does not degrade the diversity and quality of summaries.

For code generation, we consider the task of generating a Python function given its signature (name and arguments). While general-purpose language models can generate idiomatic Python functions \citep{chen2021evaluating,austin2021program}, they may still struggle to learn some desirable properties of generated code. For instance, a Python function generated by GPT-Neo will compile only 40\% of the time and will contain on average 4 violations of PEP8. We show that using CDPG to approximate a conditional EBM expressing corresponding constraints improves both compilability and PEP8-compliance without hurting the diversity of generated Python code or leading to degeneration~\citep{degeneration_HoltzmanBDFC20}.

The contributions of this paper are as follows:
\vspace{-5px}
\begin{enumerate}
    \itemsep0em 
    \item We introduce a formal framework for representing control objectives for \emph{conditional} generative models as \emph{conditional} EBMs while alleviating \emph{catastrophic forgetting},
    \item We design CDPG, an extension of DPG suitable for approximating conditional EBMs,
    \item We evaluate CDPG on three control objectives across three tasks: machine translation with number format constraints, summarization constrained to be factually correct, code generation constrained to generate compilable Python functions and code generation constrained to respect PEP8.
\end{enumerate}

Code accompanying the paper will be available at \url{https://github.com/naver/gdc}.

\section{Method}
\paragraph{Unconditional EBMs}
A standard, ``unconditional'' EBM is a function $P$ from a (discrete, i.e. finite or countable) space $X$ to the non-negative reals, such that the partition function $Z \doteq \sum_{x\in X} P(x)$ is strictly positive and finite. We will denote by lowercase $p$ the normalized distribution over $X$ associated with $P$, namely $p(x) \doteq P(x)/Z$. 
\citet{khalifa_2021} show that the problem of fine-tuning a pretrained model $a(x)$ to satisfy a control condition $b(x)=1 \ \forall x \in X$, where $b(x)$ is a binary scorer for a desired feature, while minimizing the divergence to the original a(x) has a unique solution given by the probability distribution $p$ associated with the EBM

\begin{equation}
    P(x) = a(x)b(x).
\end{equation}

In information-geometric terms, $p$ is \ptmd{the} I-projection of $a$ onto the manifold of all distributions satisfying \ptmd{the} constraint given by $b$~\citep{csizarShields2004}.%

\paragraph{Conditional EBMs}

Let us now consider a discrete, potentially infinite set $C$ of conditions $c$. Formally, $\mathcal{P}$, a \emph{conditional} EBM over $C$, is defined as a function from $C$ to the set of unconditional EBMs over $X$, in other words, a function that maps each $c\in C$ to an unconditional EBM $P_c$ over $X$:
\begin{align}
    \mathcal{P} &: c \mapsto P_c, \\
    P_c &: x \mapsto \mathbb{R}_{+}.
\end{align}
We denote by $Z_c$ the partition function of $P_c(x)$, namely: $Z_c \doteq \sum_{x\in X} P_c(x)$, and by $p_c(x)$ the normalized version of $P_c(x)$, namely: $p_c(x) \doteq P_c(x)/Z_c$.%

\paragraph{Representing constraints as conditional EBMs}

The problem of fine-tuning a pretrained model $a(x|c)$ to satisfy a control objective (e.g. generating factually correct summaries) can be seen as a constraint satisfaction problem: finding a model $p_c(x)$ that meets the demands of the control objective but at the same time stays as close as possible to the original pretrained model $a(x|c)$. We represent such an optimal model as an unconditional EBM $P_c(x)$. A control objective can be defined in terms of a
binary scorer $b(x,c)$ such that $b(x,c) = 1$ if a sample $(c,x)$ satisfies a constraint given by a control objective (e.g. $x$ is factually correct with respect to $c$) and $b(x,c) = 0$ otherwise. Let us consider a set of contexts $C$. For each $c \in C$, we can frame the problem of finding the unique model $p_c(x)$ such that (i) $b(x,c) = 1$  for all samples $x \sim p_c(x)$, and (ii) $p_c(\cdot)$ has minimal KL divergence from $a(\cdot|c)$ as an instance of the unconditional case already considered by \citet{khalifa_2021}. Following our example, $p_c$ could be a distribution over factually correct summaries of $c$ as similar as possible to a distribution over summaries which the original model $a$ would produce for a document $c$. Therefore, $p_c$ can be represented as an unconditional EBM $P_c(x)$ of the following form:\footnote{\citet{khalifa_2021} provide a more general, exponential family form of this EBM. The product-of-experts \citep{Hinton02} form in \eqref{eq:ebm} is a special case of the exponential form, see \citep[Appendix A.2]{khalifa_2021}.}
\begin{equation} 
    \label{eq:ebm}
    P_c(x) \doteq a(x|c)b(x,c).
\end{equation}%
\vspace{-10px}
\paragraph{Approximating conditional EBMs} 

While $\mathcal{P}$ represents the target conditional model optimally reconciling distance from $a(x|c)$ and the control objective, sampling and MAP decoding for
\ptmd{$\mathcal{P}$} 
is intractable for two reasons. First, \ptmd{$\mathcal{P}$} actually represents a potentially infinite collection of unconditional models of the form $p_c(\cdot)$. Second, each of these unconditional models still cannot be easily sampled from because they do not admit an autoregressive factorisation: $b(x,c)$ is only defined for the whole sequence $x$.

The second problem was addressed by \cite{opt-rl-arxiv-2019}
and \cite{khalifa_2021} who used the distributional policy gradients (DPG) algorithm to approximate unconditional EBMs $p$ using a \emph{new} unconditional model $\pit$ trained to minimize the cross-entropy $\CE(p,\pit)$. 
Unfortunately, DPG is not directly usable for a conditional model covering many contexts $c$.

To address these problems, we instead try to find a \emph{single} \ptmd{seq2seq} model $\pit$ approximating $p$ \emph{on average} across contexts. Concretely, we minimize the expected cross-entropy between $\pit$ and multiple $p_c$'s:
\vspace{-5px}
\begin{equation}
    \label{eq:objective}
    \mathcal{L}(\theta) = \EX{c \sim \tau(c)} \CE(p_c(\cdot), \pit(\cdot|c)),
\end{equation}
where the expectation is over $\tau(c)$, a distribution over $c \in C$.%
\ftk{I was wondering whether speaking of a set $C$ isn't confusing here (it may suggest MC-DPG only fine-tunes for a pre-defined set of contexts and does not generalize outside it). Maybe here we should only speak of $\tau(c)$ in abstract, and only introduce $C_\text{train}$ and $C_\text{test}$ later as approximations of  $\tau(c)$?
\textcolor{blue}{MD: IMO, $\tau$ is \emph{not} uniform over $C$, and could be over an infinite set of conditions. The loss $\mathcal{L}$, if you want, is the true average loss that would be incurred by a user at test time. Ideally, training should then indeed draw $c$ according to the same distribution. It is only for (not fully convincing to me) practical reasons that the training is limited to a few $c$'s. Actually, my impression is that, even with the MC approach to estimating $Z_c$, you could actually draw $c$ according to $\tau$ during training, and sample \emph{two} $x$'s for this $c$; just sampling one $x$ would not work with the MC estimate for $Z_c$, but two might work (I am not sure, just a hunch). I think part of the confusion comes from wanting to impose that training $c$'s and test $c$'s be disjoint, which is contrary to general ML principles. Of course, if because of practical (artificial ?) considerations, you limit the $c$'s used at training time to a small set, it may be more convincing then to ensure that elements of your test set were not seen at training time, but this is a different point.
}} 
The gradient of this objective takes the following form:
\begin{align}
    \nabla_\theta \mathcal{L}(\theta) &= \EX{c \sim \tau(c)} \nabla_\theta\CE(p_c(\cdot), \pit(\cdot|c)) \\
    &= -\EX{c \sim \tau(c)} \EX{x \sim p_c(x)} \nabla_\theta\log \pit(x|c) \\
    &= -\EX{c \sim \tau(c)} \EX{x  \sim \pit(x|c)} \frac{p_c(x)}{\pit(x|c)}   \nabla_\theta\log \pit(x|c), \label{eq:importance-sampling}\\
    &= -\EX{c \sim \tau(c)} \EX{x \sim \pit(x|c)} \frac{P_c(x)}{Z_c\pit(x|c)}   \nabla_\theta\log \pit(x|c), \label{eq:grad-est} 
\end{align}
where in \eqref{eq:importance-sampling} we applied importance sampling from $\pit$ and \eqref{eq:grad-est} expresses $p_c$ in terms of unconditional EBM $P_c$ and its  partition function $Z_c$.\footnote{This last form of the gradients estimate bears some resemblance to policy gradients methods in reinforcement learning \citep{sutton_policy_gradients} with term $P_c(x)/Z_c\pit(x|c)$ playing the role of a pseudoreward. This similarity, however, is fairly superficial: our minimization objective \eqref{eq:objective} is expected cross-entropy (over contexts), not expected (negative) reward. See \cite{RMvsDM} for extended discussion.} 

We approximate both expectations in \eqref{eq:grad-est} by sampling.
\ftk{I don't feel I made a very good job explaining that clearly. What confused me is the distinction between conditional EBMs understood as functions and partially evaluated (fixed $c$) conditional EBMs which are actually unconditional EBM. \textcolor{blue}{MD: yes, I think it would be much clearer to consider conditional EBMs as functions, as I have noted above. Basically, I was trying to introduce a notation $\bm{P}$ for what in lambda calculus would be denoted by $\lambda c \lambda x\ P(c,x)$.}} 
Intuitively, this corresponds to building unconditional EBMs $P_c(\cdot)$ on the fly for each $c \sim \tau(c)$, computing the EBM ``score'' $P_c(x)$ for each sample from the \ptmd{seq2seq model $x \sim \pit(\cdot|c)$} and then using this score as a ``pseudoreward`` term $P_c(x)/(Z_c\pit(x|c))$ in the policy gradient estimate.

\begin{algorithm}[H]
\caption{Conditional DPG (CDPG)}
\label{algo:training-loop}
\begin{algorithmic}
\begin{small}
\Require conditional EBM $P_c(x)$, initial model $a(x|c)$
\State $\pi_\theta \gets a$
\For{each iteration}
    \State $\mathcal{B} \leftarrow \{\}$ 
    \State sample batch $\{c_1, ... , c_i, ... , c_N\}$ from $\tau(c)$
    \For{each $c_i$}
        \State sample batch $\{x_1, ... , x_j , ... , x_{M}\}$ from $\pi_\theta(x|c_{i})$
        \State $\hat{Z}_{c_i} = \frac{1}{M} \sum_{j=1}^M \frac{P_{c_i}(x_j)}{\pi_\theta(x_j|c_i)}$
        \For{each $x_j$}
            \State $\mathcal{B}$ $\stackrel{+}\leftarrow$ $(x_j,c_i,\hat{Z}_{c_i})$
        \EndFor
    \EndFor
    \For{$(x,c,\hat{Z_c})$ in $\operatorname{shuffle}(\mathcal{B})$}
        \State $\theta \gets \theta + \alpha^{(\theta)} \frac{1}{\hat{Z_c} + \epsilon}\frac{P_c(x)}{\pit(x|c)} \ \nabla_\theta \log \pi_\theta(x|c)$
    \EndFor
\EndFor
\Ensure $\pi_\theta$
\end{small}
\end{algorithmic}
\end{algorithm}

\paragraph{Estimating $Z_c$}

The one term in \eqref{eq:grad-est} that remains difficult to evaluate is the partition function $Z_c$. For a single unconditional EBM, the \citep{khalifa_2021} approach had no need to evaluate the partition function $Z$ as it just scaled gradient estimates and therefore, $Z$ could be absorbed into the learning rate. In the conditional case, $Z_c$ varies with $c$. Therefore for each $c$, we compute $Z_c$ using a batch of $M$ samples $\{x_1, \dots, x_j, \dots x_{M}\}$ from $\pi_\theta(x|c_{i})$. Then, $Z_c$ is estimated using importance sampling by reweighting samples $x_j$ by their likelihood according to $\pit(\cdot|c_i)$.

\paragraph{Training loop} We train $\pit$ by stochastic gradient descent using the gradient estimate \eqref{eq:grad-est}. At each epoch, we first sample $N$ contexts $c$ and then, for each $c$, $M$ samples $x$ from $\pit(x|c)$. We maintain a buffer $\mathcal{B}$ storing each $(c_i, x_j)$ pair along with its corresponding partition function $Z_{c_i}$. Finally, we shuffle $\mathcal{B}$ and iterate over it to perform gradient steps using \eqref{eq:grad-est} with learning rate $\alpha^{(\theta)}$. The procedure is summarized in Algorithm \ref{algo:training-loop}. $\alpha^{(\theta)}$, $N$ and $M$ are hyperparameters. Values used in experiments are reported in Tables \ref{table:code-hyperparams}-\ref{table:summarization-hyperparams} in the Appendix.

\section{Experiments}
We evaluate CDPG as well as three baselines on four control objectives across three tasks: translation, summarization and code generation. Each task is associated with $C_\text{train}$, a set of contexts $c$ used for prompting the model: these are English source sentences for translation, Python function signatures in case of code generation and source documents in case of summarization. When computing evaluation metrics, we sample contexts from a held out set $C_\text{test}$ not used for training. In addition to that, for each experiment, we measure $\E_{c\sim\tau(c)} \KL(p_c, \pit)$, the expected forward KL divergence from the optimal distribution $p_c$, as well as $\E_{c\sim\tau(c)} \KL(\pit, a)$, the expected reverse KL divergence from the original pretrained model.\footnote{See Appendix \ref{sec:appendix-metrics} for details of metrics calculations.}.

\subsection{Baselines} 


\paragraph{DPG-like ablation} We compare our algorithm with an ablation (labeled as ``DPG'' on all figures) that sets $Z_c$ in the denominator of \eqref{eq:grad-est} to a constant $Z$ which is the running mean of $P_c(x)$ across $x$'s and $c$'s. This ablation resembles the original DPG algorithm for unconditional EBMs developed by \citep{A-parshakova-etal-2019-global} and extended by \citep{khalifa_2021}. While the partition function is constant for unconditional EBMs, in conditional EBMs $Z_c$ varies with $c$. Therefore, the DPG-like ablation performs gradient updates using biased gradient estimates.

\paragraph{Reinforcement learning} The problem of fine-tuning a pretrained model to satisfy a pointwise constraint $b(x,c)$ can be posed as maximising tbe expected reward $\EX{c  \sim \tau(c)}\EX{x\sim  \pit(x|c)} R(x,c)$. We consider two instances of this approach: Reinforce \citep{Williams92Reinforce} and Ziegler \citep{Ziegler19}. For Reinforce, we simply define $R(x,c) = b(x,c)$. Ziegler prevents too large departures from $a$ by adding a KL penalty term and defining $R(x,c) = b(x,c) - \beta \KL(\pit, a)$, where $\beta$ is a hyperparameter updated using an adaptive schedule.

\subsection{Translation}
\label{experiments:translation}

\begin{figure*}[h]  
    \centering
    \includegraphics[width=\linewidth]{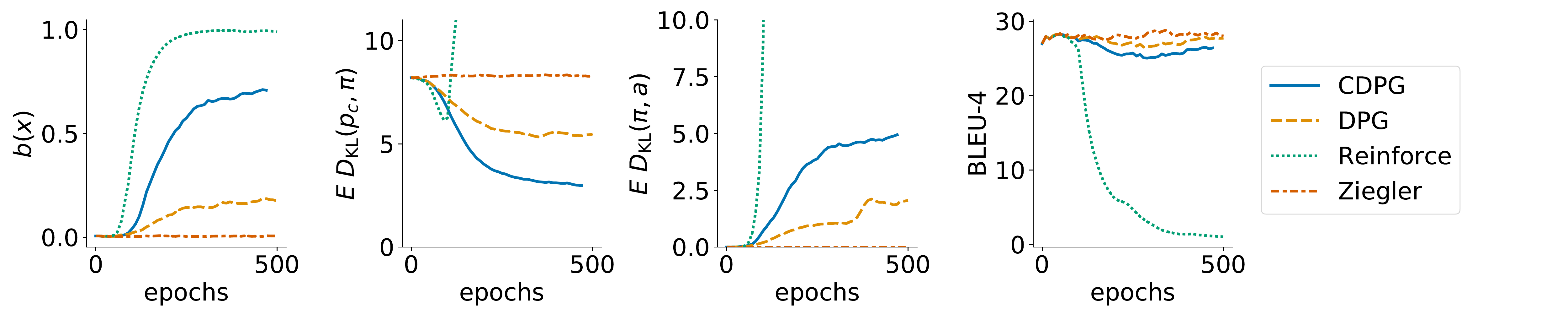} 
    \caption{\small{Translation with terminology constraint. Evaluation metrics: fraction of samples satisfying constraint $b(x)$ ($\uparrow$ better), expected $\KL(p_c,\pit)$ ($\downarrow$ better) and $\KL(\pit,a)$ ($\downarrow$ better), and BLEU-4 score ($\uparrow$ better) for models obtained from fine-tuning with conditional DPG, DPG, Ziegler and Reinforce.}}
    \label{fig:translation}
\end{figure*}

\paragraph{Dataset} For the translation task, $\tau(c)$ from Algorithm \ref{algo:training-loop} is a uniform distribution over a fixed set of English sentences. We sampled 5k English sentences containing numeral nouns from the English-French subcorpus of the Europarl dataset, version 7 \citep{koehn-2005-europarl}. Metrics are computed for generated translations of another set of 5k English sentences from the test split of Europarl. Note that neither CDPG nor the baselines utilise ground truth translations (references); we compute $b(x,c)$ based on source documents and {\it generated} translations. Ground-truth translations are only used for evaluating the BLEU score of generated translations.

\paragraph{Model} We conduct our experiments on the T5 architecture \citep{2020t5}, using the pre-trained model \texttt{t5-small} as $\pit$. During fine-tuning, we generate translations $x$ conditioned on a source sentence $c$ by pure ancestral sampling from $\pit$. For evaluation, we follow the setup described by \cite{2020t5} and use beam search decoding with beam size 4.

\paragraph{Metrics} In addition to measuring  expected $\KL(p_c,\pit)$ and $\KL(\pit,a)$, we evaluate the forgetting of T5's capabilities in terms of BLEU-4 score \cite{10.3115/1073083.1073135}, a measure of translation quality understood as overlap between generated and ground-truth translation.

\paragraph{Constraint} We implement the constraint scorer as table lookup: $b(x,c) = 1$  if for every occurrence of a given numeral noun (e.g. ``two'') in a source sentence $c$, a corresponding digit (``2'') occurs in its translation $x$. Otherwise, $b(x,c) = 0$.

\paragraph{Results}

We present the results of the translation task on Figure \ref{fig:translation}. Initial constraint satisfaction is very low: 0.006. Intuitively, it's very unlikely for T5 to translate ``two'' as ``2'', not as ``deux''. However, CDPG is able to boost that number to 0.7 and reduce the expected divergence from its target distributions $p_c$ almost twofold, outperforming DPG by a wide margin. It also outperforms Reinforce by staying closer to the original distribution $a$ and not suffering from almost any drop in BLEU-4 score. (Note that some drop is necessary for satisfying the constraint, because ground truth translations with respect to which BLEU-4 is computed almost don't satisfy the constraint themselves.) In contrast, Reinforce improves constraints satisfaction only at the cost of heavy divergence from $a$: it learns to append all the digits at the end of each translation, thus ensuring constraint satisfaction (see Appendix \ref{sec:appendix_samples_translation} for sample translations). This is reflected in a catastrophic drop in BLEU-4 score. Ziegler, on the other hand, fails to improve constraint satisfaction and stays \emph{too close} to the original distribution~$a$. 

\subsection{Summarization}

\begin{figure*}[h]  
    \centering
    \includegraphics[width=\linewidth]{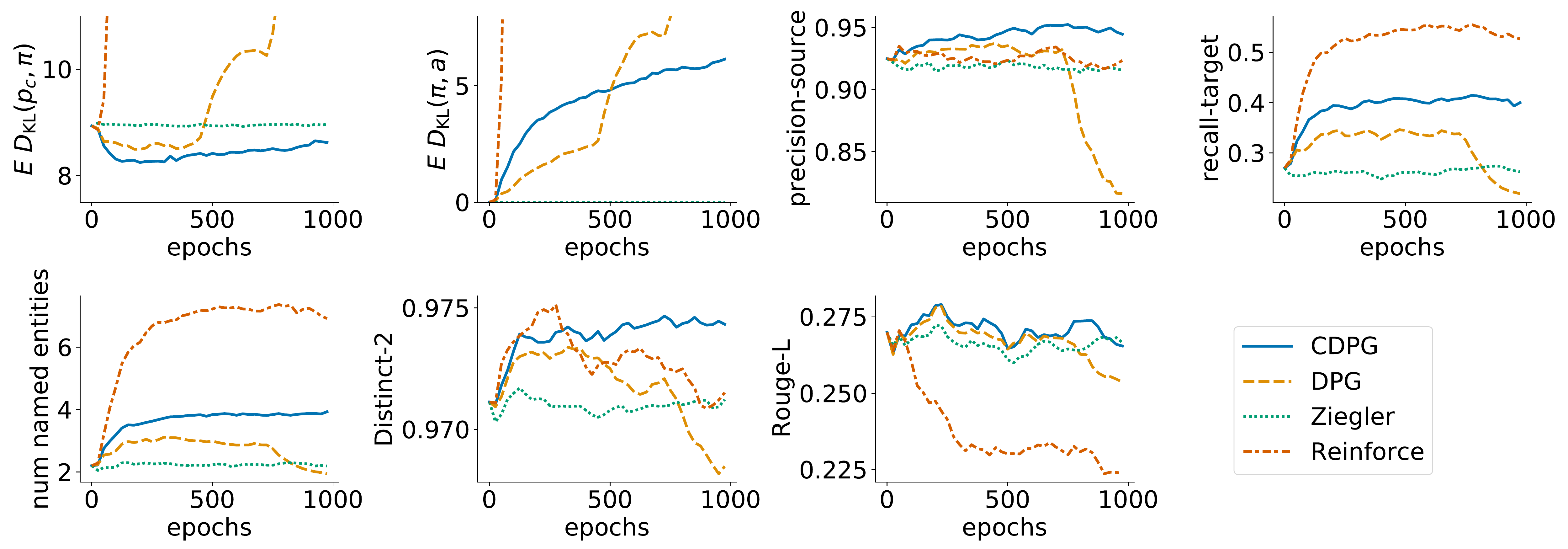} 
    \caption{\small{Summarization with factual consistency constraint. Evaluation metrics: expected $\KL(p_c,\pit)$ ($\downarrow$ better) and $\KL(\pit,a)$ ($\downarrow$ better), precision-source ($\uparrow$ better), recall-target ($\uparrow$ better), number of named entities ($\uparrow$ better), Distinct-2 ($\uparrow$ better), ROUGE-L ($\uparrow$ better) for models obtained from fine-tuning with conditional DPG, DPG, Ziegler and Reinforce.}}
    \label{fig:summarization_entities_metrics}
\end{figure*}


\paragraph{Dataset} To conduct our summarization experiments, we use the CNN/DailyMail dataset \citep{nallapati-etal-2016-abstractive} and sample 5k source documents from the train and test subsets to use for fine-tuning and evaluation, respectively.
  We use ground truth summaries only for computing reference-based evaluation metrics such as ROUGE score or recall-target. Ground truth summaries are not used in training.

\paragraph{Model} We use the same model as in the translation task (\texttt{t5-small}). For fine-tuning, we generate summaries $x$ conditioned on a source document $c$ by pure ancestral sampling from $\pit$; for evaluation, we use beam search with beam size 4.

\paragraph{Constraints} Following \cite{nan-etal-2021-entity}, we define an entity-level factual consistency constraint as a product of two constraints: there must be at least four named entities in the summary $x$ and all the named entities $x$ must have occurred in the source $c$. More formally, let $\NER(\cdot)$ denote the set of named entities found in a text and $|\cdot|$ the number of elements of a set. Then, $b(x,c) = 1$ iff $[|\NER(x)| \geq 4] \land [\NER(x) \subseteq \NER(c)]$ and $b(x,c) = 0$ otherwise.

\paragraph{Metrics} In addition to measuring expected $\KL(p_c,\pit)$ and $\KL(\pit,a)$, we evaluate the quality and factual consistency of generated summaries using the following metrics: 
\begin{enumerate}
    \itemsep0em 
    \item Precision-source \citep{nan-etal-2021-entity}, defined as $[|\NER(x) \cap \NER(x)|]/|\NER(x)|$ is the percentage of named entities in the summary that can be found in the source. Low precision-source indicates severe hallucination,
    \item Recall-target \citep{nan-etal-2021-entity}, defined as $[|\NER(x) \cap \NER(x)|]/|\NER(t)|$, is the percentage of named entities in the target summary $t$ that can be found in the generated summary $x$.
    \item Distinct-2 \citep{li-etal-2016-diversity}, a measure of text diversity in terms of the frequency of bigram repetitions within a single continuation $x$,
    \item ROUGE-L \citep{lin-2004-rouge}, a measure of summarization quality in terms of unigram overlap between the source document and ground truth summary.
\end{enumerate}
See Appendix \ref{sec:appendix-code} for on how scorers $b$ and
metrics are computed for summarization experiments.

\paragraph{Results}

We present the evolution of our 7 metrics through time on Figure \ref{fig:summarization_entities_metrics}. CDPG is the only method stably decreasing expected $\KL(p_c,\pit)$ and thus approaching (as opposed to drifting away from) optimal distributions $p_c$. This is reflected in moderate divergence from $a$ and translates into downstream metrics. Summaries generated by the fine-tuned model contain, on average, more named entities. Moreover, name entities in summaries are both more factually consistent with source (an increase in precision-source) and more relevant (an increase in recall-target). The tendency towards mentioning more factually consistent named entities increases the bigram diversity within summaries (Distinct-2) and the overall quality of generated summaries compared to ground truth (ROUGE-L). This last results might seem surprising given that CDPG did \emph{not} have access to ground truth summaries. A plausible explanation is that the original pretrained model was biased towards mentioning too few factually correct entities, at least compared to ground truth summaries. Satisfying the factual consistency constraint reduced this bias.

Baseline approaches fall short of achieving similar results. The DPG-like ablation, the closest contender, still leaves a significant gap in terms of \emph{all} metrics and is far less stable than CDPG (e.g. its $\KL(p_c,\pit)$ starts to diverge again after around 500 epochs). Ziegler stays extremely close to the original model $a$, failing to improve its shortcomings. In contrast, Reinforce heavily departs from $a$ pushing it to mention a large number of named entities. This results in artificially inflated recall-target but no increase in precision-source and a \emph{decrease} in ROUGE-L. The additional named entities frequently are frequently irrelevant (i.e. not mentioned in ground truth summaries) or simply hallucinated. See Tables \ref{tab:sum_0_samples}-\ref{tab:sum_4_samples} in the Appendix for randomly chosen summaries of documents in the test set.

\begin{figure*}[h]  
    \centering
    \begin{subfigure}[t]{\textwidth}
    \vskip 0pt
    \centering
    \includegraphics[width=\linewidth]{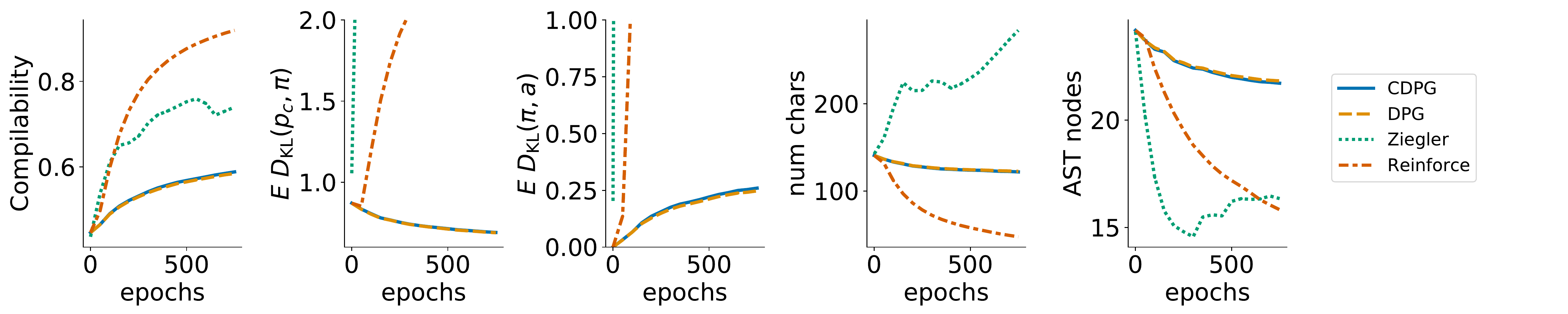}
    \caption{\small{Compilability constraint}}
    \end{subfigure}
    \begin{subfigure}[t]{\textwidth}
    \vskip 0pt
    \centering
    \includegraphics[width=0.97\linewidth]{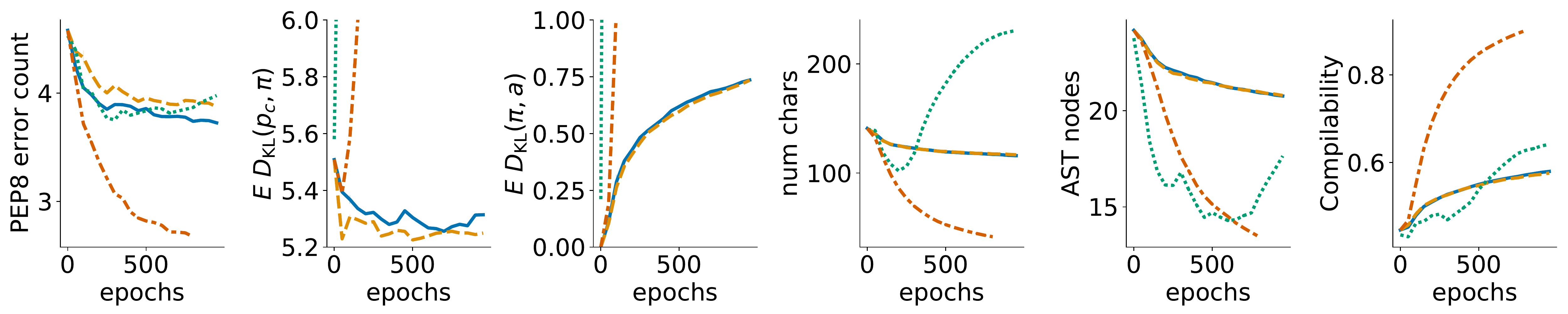}
    \caption{\small{PEP8 constraint}}
    \end{subfigure}
    \caption{\small{Code generation with compilability (a) and PEP8 (b) constraint. Evaluation metrics: compilability ($\uparrow$ better), number of PEP8 errors ($\downarrow$ better), expected $\KL(p_c,\pit)$ ($\downarrow$ better) and $\KL(\pit,a)$ ($\downarrow$ better),  number of characters, AST node count ($\uparrow$ better) for models obtained from fine-tuning with CDPG, DPG, Ziegler and Reinforce.}}
    \label{fig:code_metrics}
\end{figure*}

\begin{figure*}[h]  
    \centering
    \begin{subfigure}[t]{\textwidth}
    \vskip 0pt
    \centering
    \includegraphics[width=\linewidth]{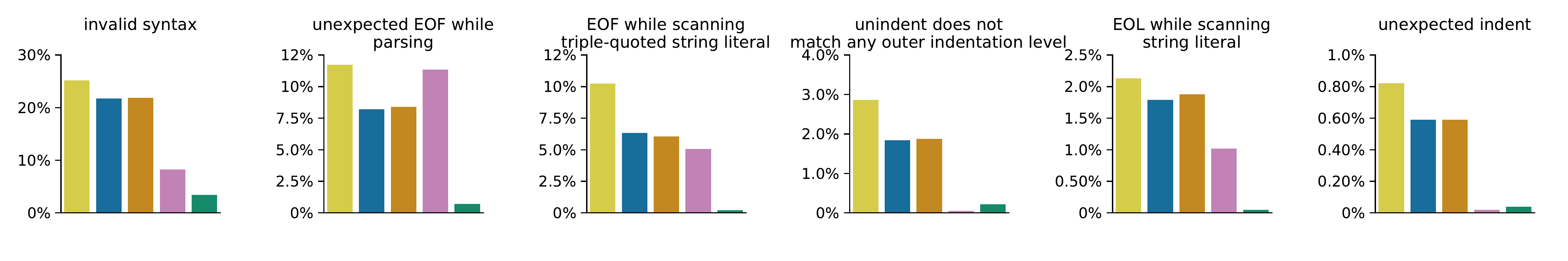}
    \caption{\small{Relative frequencies of compilation errors}}
    \end{subfigure}
    \begin{subfigure}[t]{\textwidth}
    \vskip 0pt
    \centering
    \includegraphics[width=\linewidth]{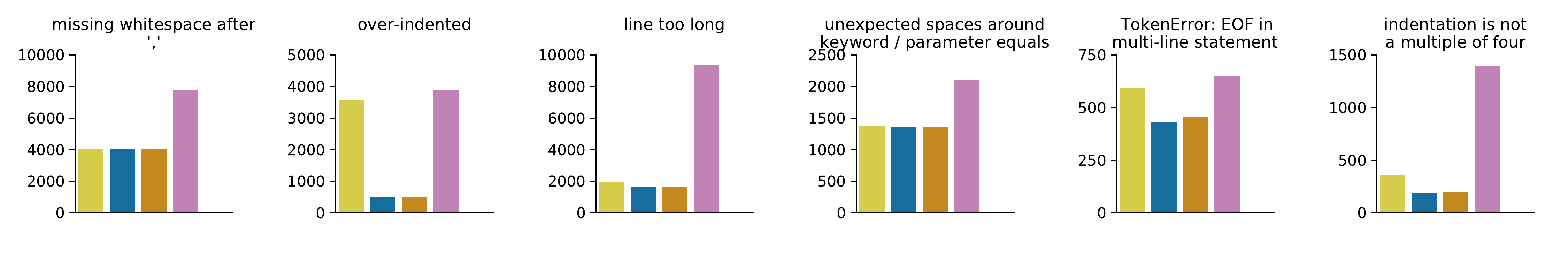}
    \caption{\small{Absolute frequencies of PEP8 violations}}
    \end{subfigure}
        \begin{subfigure}[t]{\textwidth}
    \vskip 0pt
    \centering
    \includegraphics[width=\linewidth]{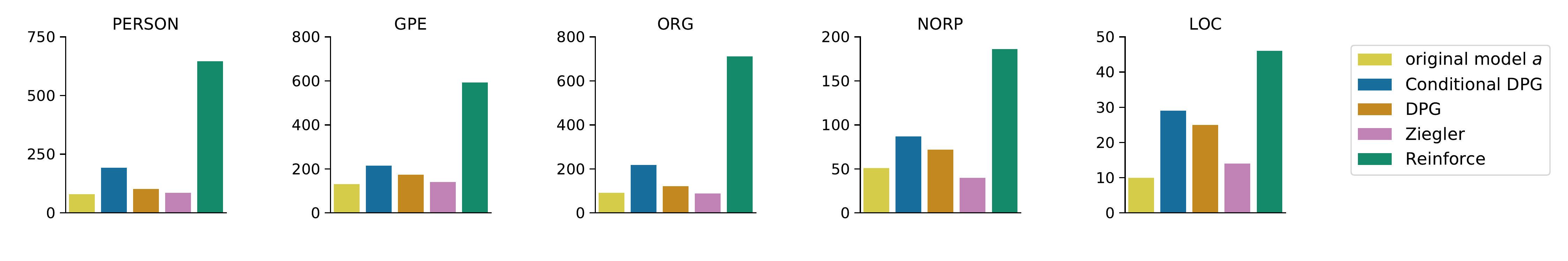}
    \caption{\small{Absolute frequencies of named entities}}
    \end{subfigure}
    \caption{\small{Relative frequencies of most common compilation errors (a), absolute frequencies of most common PEP8 violations (b) and absolute frequencies of named entities (c) in a batch of 10280 samples from the original model $a$ as well as models obtained from fine-tuning with CDPG, DPG, Ziegler and Reinforce.}}
    \label{fig:errors}
\end{figure*}

\begin{figure*}[h]  
    \centering
    \begin{subfigure}[t]{.24\textwidth}
    \vskip 0pt
    \centering
    \includegraphics[width=\linewidth]{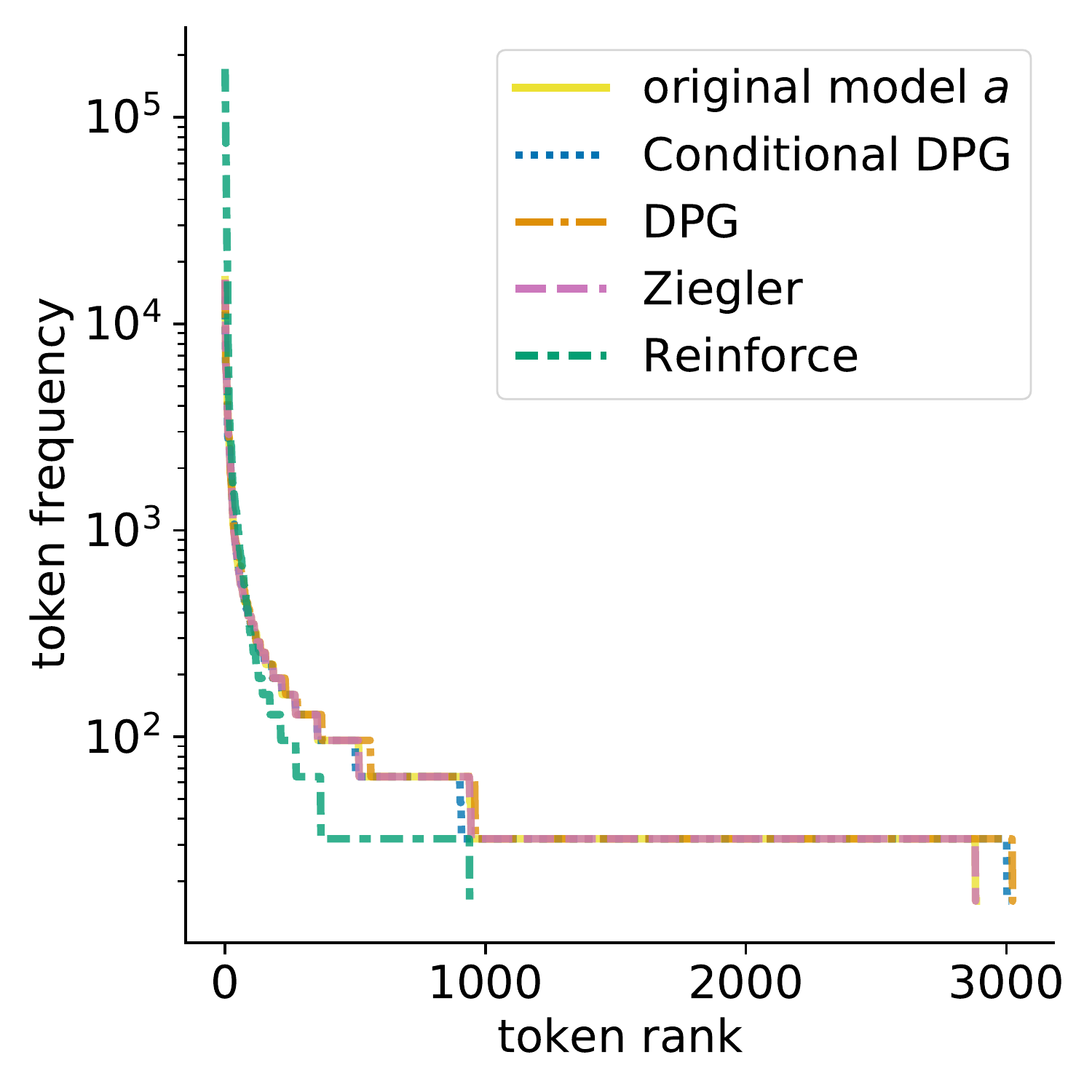}
    \caption{\small{Translation with terminology constraint}}
    \end{subfigure}
    \begin{subfigure}[t]{.24\textwidth}
    \vskip 0pt
    \centering
    \includegraphics[width=\linewidth]{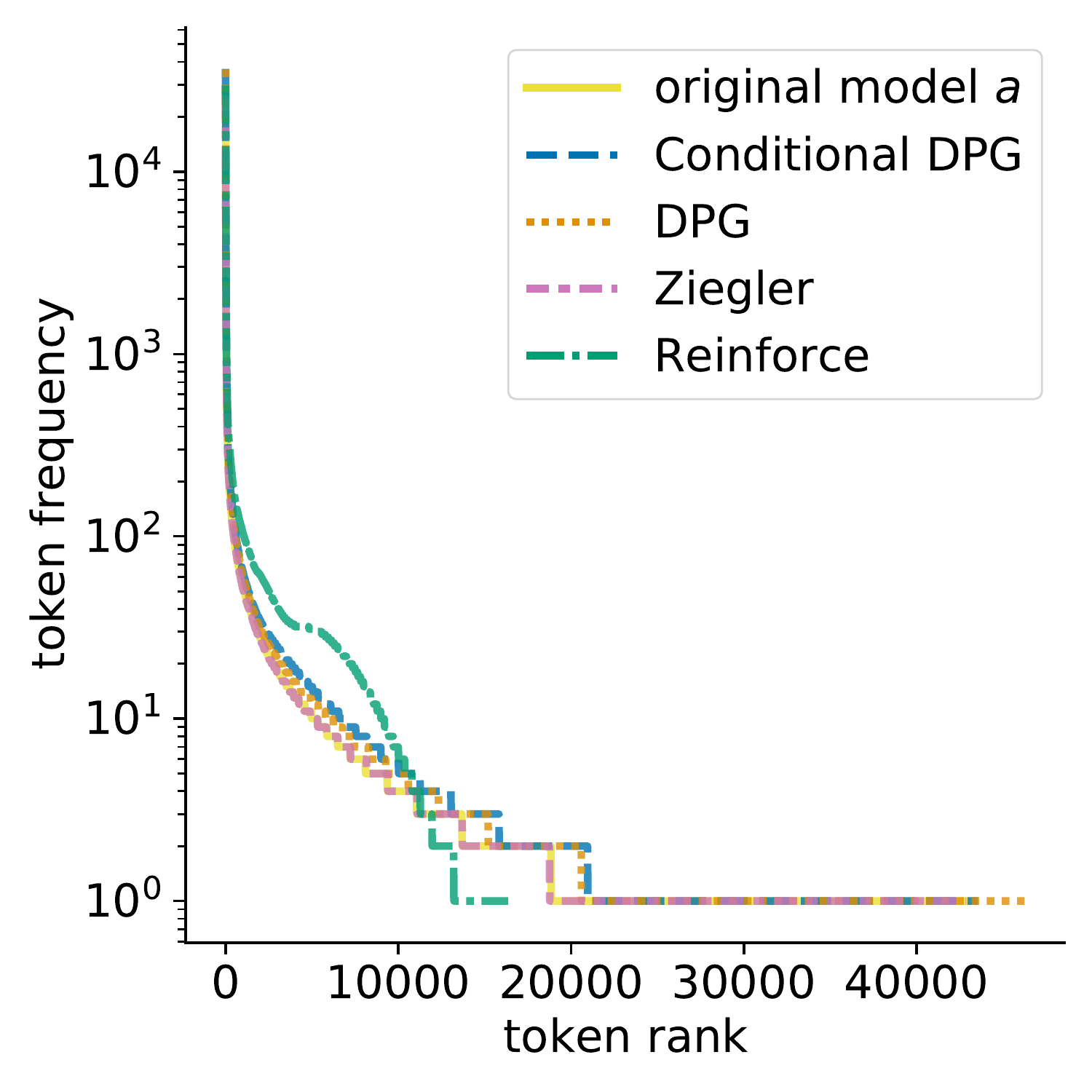}
    \caption{\small{Summarization with factual consistency constraint}}
    \end{subfigure}
    \begin{subfigure}[t]{.24\textwidth}
    \vskip 0pt
    \centering
    \includegraphics[width=\linewidth]{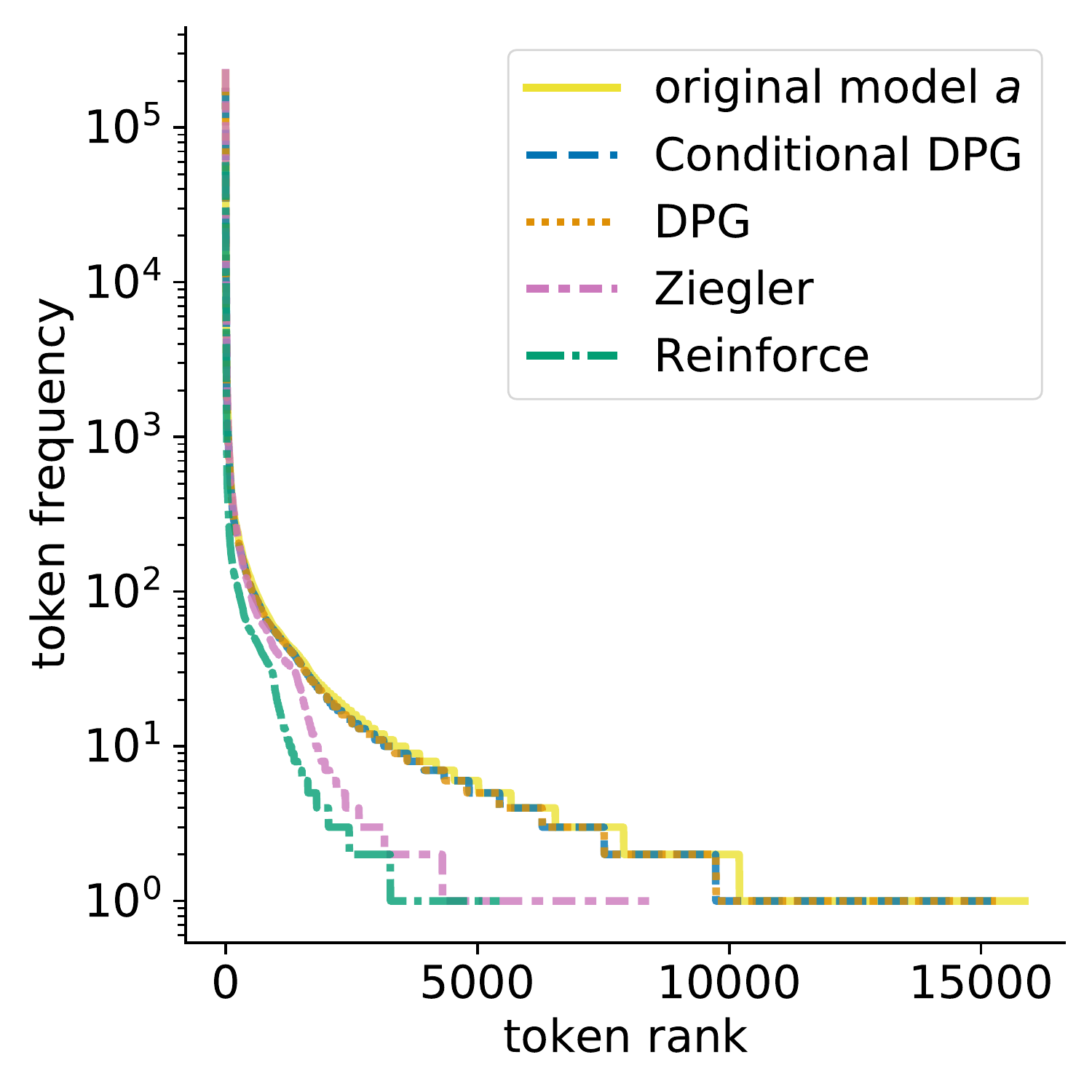}
    \caption{\small{Code generation with compilability constraint}}
    \end{subfigure}
    \begin{subfigure}[t]{.24\textwidth}
    \vskip 0pt
    \centering
    \includegraphics[width=\linewidth]{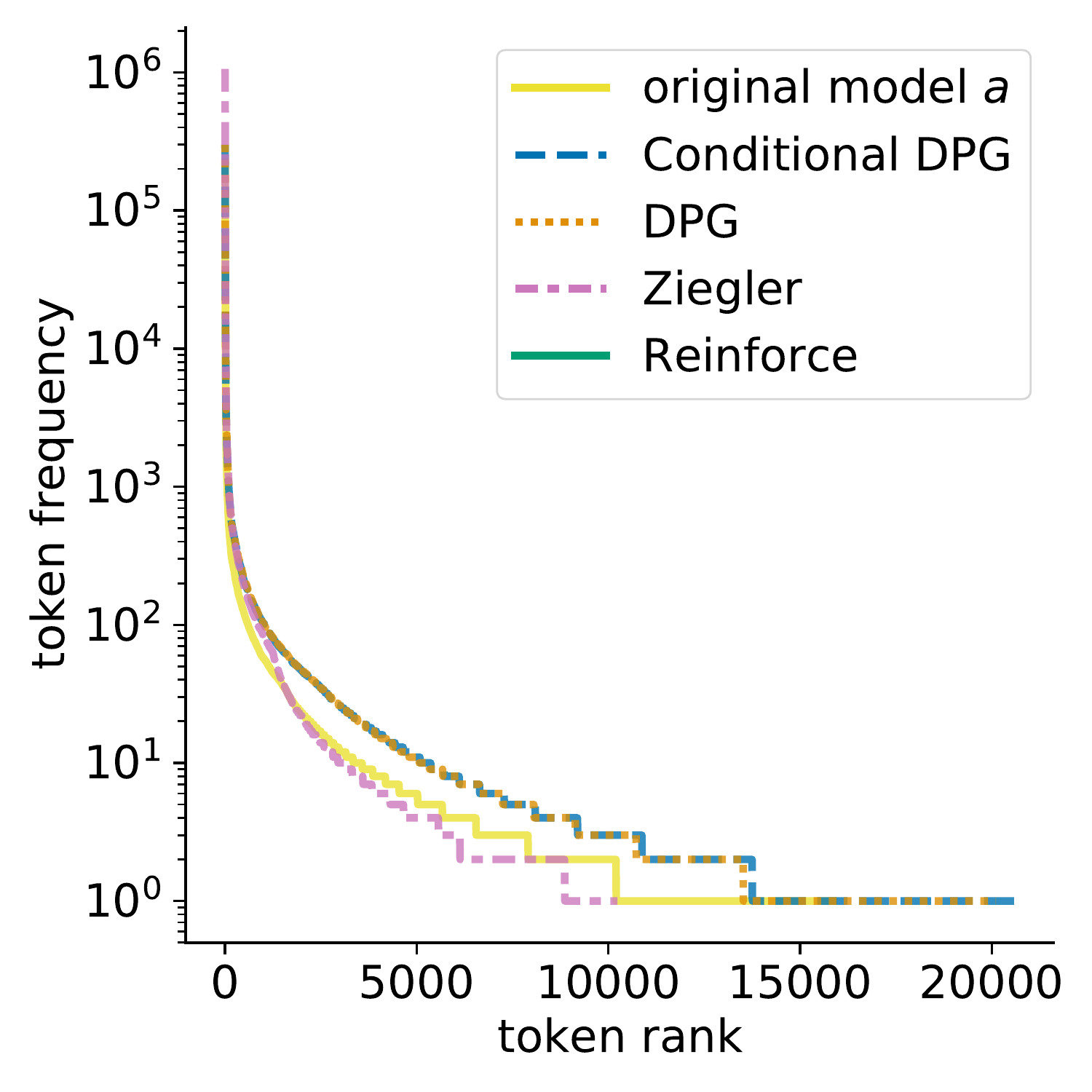}
    \caption{\small{Code generation with PEP8 constraint}}
    \end{subfigure}
    \caption{\small{Token frequency against token rank computed for tokens in 10280 samples from $a$, Conditional DPG and baselines. Longer tails imply more diverse samples.}}
    \label{fig:zipf}
\end{figure*}

\subsection{Code generation}

\paragraph{Dataset}

For code generation experiments, we condition a language model on Python function signatures (both of methods and standalone functions) extracted from the Python150 dataset which consists of Python source code  obtained from GitHub \citep{Raychev2016}. We use the code provided by  \cite{roziere2020unsupervised} for function extraction and randomly choose 5k functions for $C_\text{train}$ and 5k for $C_\text{test}$. $\tau(c)$ is a uniform distribution over these signatures. Note that neither in fine-tuning nor in evaluation do we use ground truth function bodies.

\paragraph{Model} We conduct experiments using GPT-Neo \citep{gpt-neo}: an off-the-shelf, freely available autoregressive language model mirroring the GPT-3 architecture \citep{gpt3}. GPT-Neo's training set included 85 GiB of source code from GitHub which endowed it with some code completion abilities \citep{gao2020pile}. We use the \texttt{gpt-neo-125} variant available on Huggingface Transformers \citep{huggingface}. During both fine-tuning and evaluation we generate function bodies by conditioning on signatures using pure ancestral sampling.
\paragraph{Constraints}

For experiments with compilability control condition, we check compilability of a Python function declaration obtained by concatenating $[c,x]$ and trying to execute it. $b(x,c) = 1$ if the Python interpreter raises an exception and 0 otherwise. See Appendix \ref{sec:appendix-code} for more details.

For experiments with PEP8-compliance control condition, we check whether a function declaration given by $[c,x]$ violates PEP8 \citep{pep8}, the style guide for Python, by running \texttt{pycodestyle}, an off-the-shelf linter (static code analysis tool).\footnote{\url{https://github.com/PyCQA/pycodestyle}} $b(x,c) = 1$ if the number of PEP8 violations found by \texttt{pycodestyle} is 0, otherwise $b(x,c) = 0$.

\paragraph{Metrics}
We evaluate the quality of generated Python functions using the following metrics:
\begin{enumerate}
    \itemsep0em 
    \item PEP8 error count, the average  number of violations of PEP8,
    \item Compilability, the fraction of samples $[c,x]$ that compile,
    \item The average number of characters in $[c,x]$ (after detokenization),
    \item The average number of nodes in an abstract syntax tree (AST) of sequences that compile. Intuitively, this metric indicates the logical (as opposed to surface) complexity of generated programs.
\end{enumerate}
For more details on how scorers $b$ and metrics are implemented, see Appendix \ref{sec:appendix-code}.

\paragraph{Results}

We present the evolution of metrics through time on Figure \ref{fig:code_metrics}. CDPG was able to increase the fraction of compilable functions from around 40\% to around 65\% and decrease the average number of PEP8 violations. Incidentally, the PEP8 control objective also leads to an increase in compilability because many PEP8 violations are also compilation errors. Here we note similarly to the results of previous experiments that, CDPG and its DPG-like ablation are the only methods actually approaching optimal distributions $p_c$ and diverging moderately from $a$. This allows them to maintain the original statistics of $a$: length and the number of nodes in AST trees of generated functions. In contrast, Reinforce learns to generate shorter functions (having less opportunity for mistakes) and Ziegler produces heavily degenerated samples \citep{degeneration_HoltzmanBDFC20}: syntactically simple functions with severe repetitions. This is reflected in an increase in length and a decrease in AST nodes count. See Tables \ref{tab:code_0_samples}-\ref{tab:code_2_samples} and Tables \ref{tab:code_pep8_0_samples}-\ref{tab:code_pep8_2_samples} for randomly chosen samples from the compilability and PEP8 experiments, respectively.

Note that the performance gap between CDPG and its DPG-like ablation is much closer for code generation (especially with compilability control objective) than for summarization. This can be accounted for by the normalized standard deviation of partition functions $Z_c$ for EBMs $P_c$ in the range of conditional EBMs $\mathcal{P}$ for each control objective. For code generation, this standard deviation is lower meaning that $Z_c$ in \eqref{eq:grad-est} is better approximated by a constant which can be absorbed into the learning rate $\alpha^{(\theta)}$. For summarization, this variance is higher, therefore ignoring the $Z_c$ term incurs higher bias which translates into worse performance. See Appendix \ref{sec:appendix_nstd_zc} for a comparison.
\subsection{Qualitative analysis}

In the previous sections, we showed how CDPG is able to fine-tune a pretrained model $a$ to satisfy certain constraints without destroying $a$'s capabilities. Here we attempt to gain a better understanding of how different fine-tuning approaches affect the distributions of final models. On Figure \ref{fig:errors} we present frequencies of errors (for the code generation task) and named entities (for the summarisation task) obtained from fine-tuned models. While errors and named entities differ significantly in their frequency, CDPG consistently decreases frequencies of these errors and consistently increases the frequencies of all kinds of named entities, including the long tail of rare ones.

To compare lexical diversity of samples obtained from fine-tuned models (for all four tasks), we plot the frequency of each token (the number of times it occurs) and its rank (its index in a sorted list of tokens) in Figure~\ref{fig:zipf}. CDPG and its DPG-like ablation are able to closely match original token frequencies while Ziegler and Reinforce tend to have shorter tails of rare tokens. 

\section{Related work}
\paragraph{Injecting prior information in machine translation} 
The ``Statistical Machine Translation'' paradigm \citep{Koehn2010}, which was dominant before the deep learning revolution in NLP, heavily exploited log-linear models over predefined features of
translation pairs, but without the ability to learn representations typical of neural approaches. Building over such prior work,  \citet{Zhang2017PriorKI} propose to inject a regularization term over a neural translation model which asks for the posterior distribution to be close to a log-linear model using predefined features. This approach is similar to ours in that it allows to incorporate arbitrary features into the posterior distribution. In contrast to our work, the conditional model must be trained jointly with the log-linear one, instead of allowing to control an existing pre-trained model towards directly satisfying constraints.
The task here explored is in the spirit of machine translation with terminological constraints~\citep{chatterjee2017guiding, hasler2018neural, dinu2019training, ibn2021findings}. Some approaches to tackle it include constrained decoding~\citep{chatterjee2017guiding,hasler2018neural}, adding the desired terminology as part of the source context~\citep{dinu2019training}, among others. Unlike the here-presented approach, these approaches are specific to this task only and will not generalize to arbitrary constraints.

\paragraph{Reducing hallucinations in summarization} Neural abstractive summarization is highly prone to hallucinate content in the summary that is unfaithful to the source document.  \citet{maynez-etal-2020-faithfulness} found that hallucinations occur in more than 70\% of single-sentence summaries and most of these are {\it extrinsic hallucinations}: adding information not directly inferable from the input document. Therefore, a substantial effort was devoted to to improving factual consistency of abstractive summarization. Some notable attempts include reranking summaries based on their correctness predicted by entailment classifiers \citep{falke-etal-2019-ranking} or fine-tuning using RL with a reward derived from an entailment classifier \citep{pasunuru-bansal-2018-multi}. The notion of {\it entity-level} factual consistency -- a property such that all named entities in the summary are actually mentioned in the source document -- was introduced by \citet{nan-etal-2021-entity} as one way of operationalizing the notion of extrinsic hallucinations.

\paragraph{Controllable code generation}

Generating source code is an established application of language models \citep{Nguyen2013,Raychev2014,Karpathy2015VisualizingAU,Bielik2016} that since recently has enjoyed renewed interest \citep{codexglue,chen2021evaluating,austin2021program}. The task is formulated both as unconditional generation (with applications in code completion, e.g. Codex \citep{codexglue} or GitHub Copilot\footnote{\url{https://copilot.github.com}}) and as conditional generation (e.g. program synthesis or generating a program satisfying a given input-output specification, e.g. \citep{austin2021program}). Our task of function generation can be seen as a simplified program synthesis with the specification given by function signature (a name of a function and a list of arguments). Previous work found compilability errors to be a signification failure mode of neural code generation \citep{roziere2020unsupervised}. Previous attempts at improving compilability of generated code include \cite{Maddison2014}, who augment neural probabilistic context free grammars with semantic constraints and use them for unconditional generation or \cite{zhong2017seq2sql}, who used policy gradients to train a model translating natural language questions to corresponding SQL queries and -- in addition for rewarding for query execution results -- added a penalty for syntactically invalid queries. Most in line with our work, \citet{korbak2021energybased} used DPG to improve compilability of unconditional language models for code.

\section{Conclusion}
We presented CDPG, a principled approach to fine-tuning conditional language models to satisfy arbitrary constraints. In contrast with other methods, CDPG does not require ground truth training data and is able to shift model distribution in a minimally invasive way. In consequence, models fine-tuned with CDPG share desired characteristics, such as improved factual consistency or compilability, with the fluency and diversity of the original model.

Future work could evaluate CDPG on other tasks --- such as dialogue --- as well as explore other control objectives such as constraining the semantics (as opposed to syntax) of generated Python functions. Another future direction consists in extending CDPG to approximate conditional analogues of the more general, \emph{exponential-form} \citep{khalifa_2021} EBMs which can represent \emph{distributional} constraints, namely, desired expected values for certain features of generated samples.
\bibliography{references,additional_references}
\bibliographystyle{icml2022}

\newpage
\appendix
\onecolumn


\section{Details of metric and score calculation}\label{sec:appendix-metrics}

\subsection{KL divergences}\label{sec:appendix-kls}

Calculation of metrics relative to $p_c$'s, such as $\EX{c \sim \tau(c)} \KL(p_c,\pit)$, requires estimating  $Z_c$'s. This is done using importance sampling from $\pit$ in a manner analogous to the training loop (Algorithm \ref{algo:training-loop}). Then, expected KL can be simplified to the following form:
\begin{align}
\E_{c \sim \pi(c)} \KL \Big[ p_c(x) \ ||\  \pit(x|c) \Big] &= \E_{c \sim \pi(c)}  \sum_x p_c(x) \log \frac{p_c(x)}{\pit(x|c)} \\
&= \E_{c \sim \pi(c)} \sum_x p(x|c) \log \frac{P_c(x)}{Z_c \pit(x|c)} \\
&= \E_{c \sim \pi(c)} \Big[ -\log Z_c + \sum_x p(x|c) \log \frac{P_c(x)}{\pit(x|c)} \Big] \\
&= \E_{c \sim \pi(c)} \Big[ -\log Z_c + \sum_x \pit(x|c) \frac{P_c(x)}{\pit(x|c)} \log \frac{P_c(x)}{\pit(x|c)} \Big]\\
&= \E_{c \sim \pi(c)} \Big[-\log Z_c + \frac{1}{Z_c} \mathbb{E}_{x\sim \pit(x|c)} \frac{P_c(x)}{\pit(x|c)} \log \frac{P_c(x)}{\pit(x|c)} \Big].
\end{align}

A small $\epsilon$ is added to $Z_c$ for stability. We approximate both expectations (over $\tau$ and $\pit$) using importance sampling. For a complete procedure, see Algorithm \ref{algo:kl}. 

$\EX{c \sim \tau(c)} \KL(\pit,  a)$ is computed in a simpler manner as it doesn't require estimating $Z_c$'s and we can directly sample from $\pit$. It boils down to sampling a batch of $N$ contexts $c_i$, a batch of $M$ samples $x_j$ from $\pit(x|c_i)$ for each $c_i$ and evaluating:
\begin{equation}
    \EX{c \sim \tau(c)} \KL(\pit, a) \approx \frac{1}{NM} 
\sum_{i=1}^N \sum_{j=1}^M \frac{\pit(x_j|c_i)}{a(x_j|c_i)}.
\end{equation}

To avoid bias, when computing KL divergences we always sample from $\pit$ using pure ancestral sampling (as opposed to top $p$ sampling or beam search decoding).

\begin{algorithm}[H]
\caption{Estimating $\EX{c \sim \tau(c)} \KL(p_c,\pit)$}
\label{algo:kl}
\begin{algorithmic}[1]
\Require a distribution over queries $\tau(c)$
\Require conditional model $\pit$
\Require $N$, number of contexts
\Require $M$, number of samples for each context
\State sample batch $\{c_1, ... , c_i, ... , c_N\}$ from $\tau(c)$
\For{$i \in \{1, \dots, N\}$}
    \State sample batch $\{x_1, ... , x_j , ... , x_{M}\}$ from $\pi_\theta(x|c_{i})$
        \State $\hat{Z}_{c_i} = \frac{1}{M} \sum_{j=1}^M \frac{P_{c_i}(x_j)}{\pi_\theta(x_j|c_i)}$
\EndFor 
\State $\KL(p,\pit) = \frac{1}{NM} 
\sum_{i=1}^N \sum_{j=1}^M \Big[\frac{1}{\hat{Z}_{c_i} + \epsilon} \frac{P_{c_i}(x_j)}{\pit(x_j|c_i)} \Big[ -\log \hat{Z}_{c_i} + \log \frac{P_{c_i}(x_j)}{\pit(x_j|c_i)} \Big] \Big]$
\Ensure An estimate of $\EX{c \sim \tau(c)} \KL(p_c,\pit)$
\end{algorithmic}
\end{algorithm}

\subsection{Translation}

We implement the scorer for number normalization as a lookup table mapping a numeral noun (e.g. ``one'') to a digit (``1''). Digits range from 1 to 9. A constraint is satisfied if for every occurrence of a given numeral noun in source sentence $x$, a corresponding digit occurs in its translation $x$.

To compute BLEU-4 score, we use the SacreBLEU implementation \citep{post-2018-call}.

\subsection{Summarization}\label{sec:appendix-summarization}



Following \cite{nan-etal-2021-entity}, we implement $\NER(\cdot)$ as using a pretrained SpaCy \citep{spacy} named entity recognizer. We use the \texttt{en\_core\_web\_sm} model and restrict the named entities we extract to the following categories: \texttt{PERSON}, \texttt{FAC} (buildings, airports, highways, bridges, etc.), \texttt{GPE} (geopolitical entities: countries, cities, etc.), \texttt{ORG} (companies, agencies, institutions, etc.), \texttt{NORP} (nationalities or religious or political groups), \texttt{LOC} (Non-GPE locations: mountain ranges, bodies of water, etc.), \texttt{EVENT} (named hurricanes, battles, wars, sports events, etc.). Also following \cite{nan-etal-2021-entity}, we ignore entities such as date, time and numerals due to large variation in their representation in documents.

\subsection{Code generation}\label{sec:appendix-code}

\paragraph{Compilability} To check for compilability, we call the \texttt{compile\_command} function from the \texttt{codeop} module of Python Standard Library\footnote{\url{https://docs.python.org/3/library/codeop.html}} with a sequence obtained by string concatenation $[c, x]$ as argument. We then check if \texttt{compile\_command} returns a \texttt{code} object. The only postprocessing we apply is removing any characters from $x$ after the end of function declaration (with function end defined in terms of indentation) as we are concerned specifically with function generation. 
\texttt{codeop.compile\_command} is the implementation that Python interactive interpreters use in read-eval-print loop (REPL) to determine whether a string is a valid Python code. The method tries to compile a string of Python code and raise and exception if compilation fails, for instance 
a \texttt{SyntaxError} for invalid Python syntax and \texttt{ValueError} or \texttt{OverflowError} if there is an invalid literal. Note that our notion of compilability is concerned only with syntactic correctness as Python interpreter does not execute the body of a function at function declaration time.

\paragraph{PEP8} To compute the number of PEP8 violations triggered by a sequence $[c,x]$, we  run pycodestyle,\footnote{\url{https://github.com/PyCQA/pycodestyle}} a Python linter (static code analysis tool) and report the number of violations it reports.

\paragraph{AST node count} Finally, to compute AST node count, the average number of nodes in an abstract syntax trees (ASTs) of generated functions, we consider only samples $[c,x]$ that compile. They are parsed to their corresponding ASTs using the \texttt{ast} module from Python Standard Library.\footnote{\url{https://docs.python.org/3/library/ast.html}} 



\subsection{Normalized standard deviations for $Z_c$ across tasks}\label{sec:appendix_nstd_zc}

\begin{figure*}[h]  
    \centering
    \includegraphics[width=0.5\linewidth]{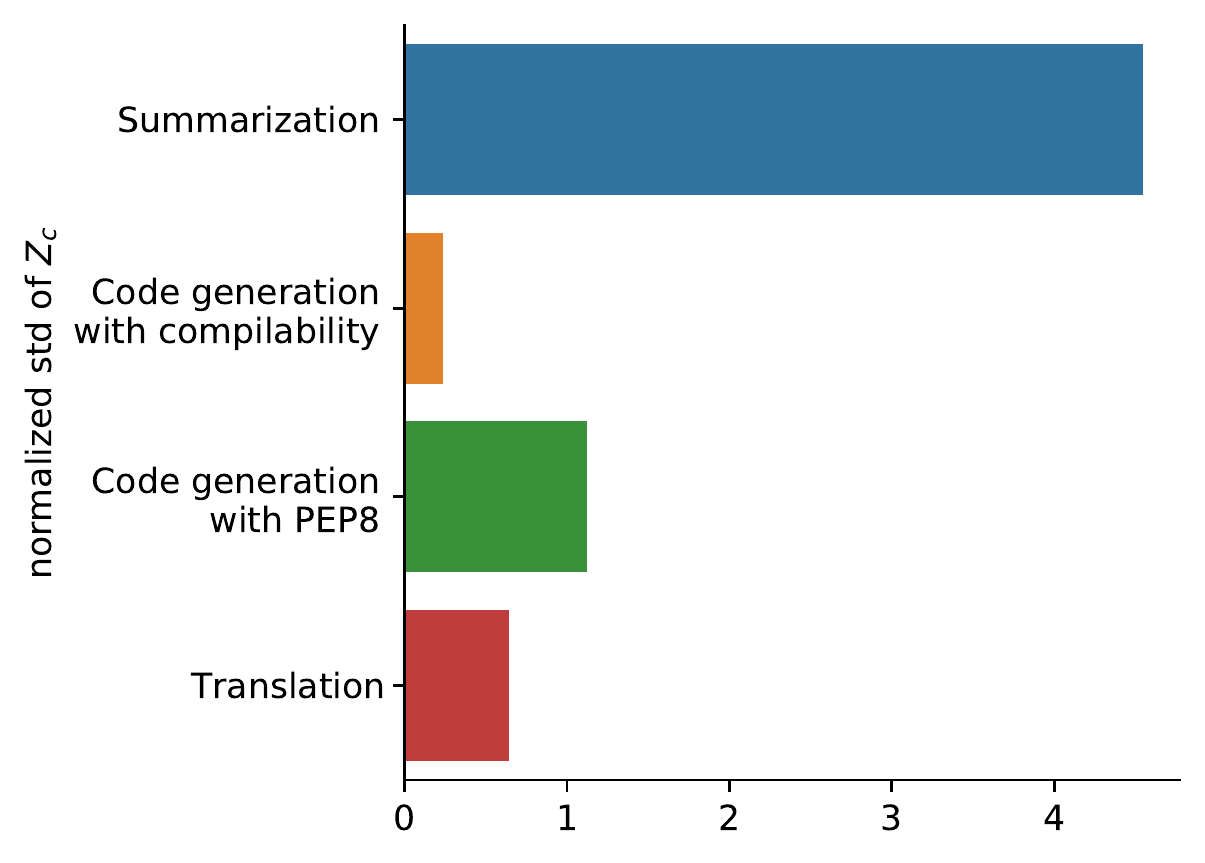}
    \caption{\small{%
    Normalized standard deviations for $Z_c$'s associated with unconditional EBMs $P_c$ in the codomain of conditional EBMs $\mathcal{P}$ defined for three control objectives: summarization with factual consistency constraint, code generation with compilability constraint and code generation with PEP8 constraint.}}
    \label{fig:nstd_zc}
\end{figure*}

See Figure \ref{fig:nstd_zc} of normalized standard deviations of $Z_c$ across tasks. Here normalized standard deviations is defined as $\text{std}(Z_c)/\text{avg}(Z_c)$, where
\begin{align}
    \text{avg}(Z_c) &= \frac{1}{N} \sum_{i=1}^N Z_{c_i}, \\
    \text{std}(Z_c) &= \sqrt{\frac{1}{N} \sum_{i=1}^N \Big(Z_{c_i} - \text{avg}(Z_c)\Big)^2}.
\end{align}
Lower normalized standard deviation for a task explains closer performance gap between CDPG and DPG for that task on Figures \ref{fig:summarization_entities_metrics}-\ref{fig:code_metrics}.

\section{Hyperparmeters and implementation details}

We implemented all models using PyTorch~\citep{pytorch} and HuggingFace Transformers ~\citep{huggingface}. Each training run took approximately 5 days on 2 Nvidia V100 GPUs. For a detailed list of hyperparameter values, see Table \ref{table:code-hyperparams} and \ref{table:summarization-hyperparams}. For a description of hyperparameters specific to Ziegle see \citep{Ziegler19}.

\begin{table}[H]
    \footnotesize
    \centering
    \begin{tabular}{lll}
    \toprule
    \textbf{Hyperparameter} & \textbf{Value} & \textbf{Symbol} \\
    \toprule
    \multicolumn{3}{c}{\textbf{Common}} \\
    original model & \texttt{EleutherAI/gpt-neo-125M} & $a$ \\
    batch size & 2048 \\
    maximum sequence length & 128 tokens \\
    learning rate for $\pit$ & $1.41 \times 10^{-6}$ &  $\alpha^{(\theta)}$\\
    optimizer & Adam \citep{kingma2014adam}\\
    learning rate schedule & constant with warmup (100 epochs) \\
    total epochs & 1000 \\
    number of $c$'s for training & 5000 & $|C_\text{train}|$ \\
    number of $c$'s per batch & 32 & $N$  \\
    number of $x$'s per $c$ & 64 & $M$ \\
    \multicolumn{3}{c}{\textbf{Ziegler}} \\
    policy gradients clip range & 0.2\\
    target KL value for adaptive schedule & 6.0 \\
    initial coefficient of KL penalty & 0.2 & $\beta$ \\
    \bottomrule
    \addlinespace
    \end{tabular}
    \caption{Hyperparameters used for code generation experiments}
    \label{table:code-hyperparams}
\end{table}

\begin{table}[H]
    \footnotesize
    \centering
    \begin{tabular}{lll}
    \toprule
    \textbf{Hyperparameter} & \textbf{Value} & \textbf{Symbol} \\
    \toprule
    \multicolumn{3}{c}{\textbf{Common}} \\
    original model & \texttt{t5-small} & $a$ \\
    batch size & 1024 \\
    maximum sequence length & 200 tokens \\
    learning rate for $\pit$ & $1 \times 10^{-4}$ &  $\alpha^{(\theta)}$\\
    optimizer & Adam \citep{kingma2014adam}\\
    learning rate schedule & constant with warmup (100 epochs) \\
    total epochs & 1000 \\
    number of $c$'s for training & 5000 & $|C_\text{train}|$ \\
    number of $c$'s per batch & 32 & $N$  \\
    number of $x$'s per $c$ & 32 & $M$ \\
    \multicolumn{3}{c}{\textbf{Ziegler}} \\
    policy gradients clip range & 0.2\\
    target KL value for adaptive schedule & 6.0 \\
    initial coefficient of KL penalty & 0.2 & $\beta$ \\
    \bottomrule
    \addlinespace
    \end{tabular}
    \caption{Hyperparameters used for translation and summarization experiments}
    \label{table:summarization-hyperparams}
\end{table}


\section{Samples}\label{sec:appendix_samples}

\subsection{Translation with terminology constraint}\label{sec:appendix_samples_translation}

See Tables \ref{tab:trans_0_samples}-\ref{tab:trans_4_samples} for translations of 5 randomly chosen source sentences from our evaluation set $C_\text{test}$ generated by models fine-tuned using Conditional DPG, 3 baselines and $a$. Translations were generated with beam search.

\subsection{Summarization with entity-level factual consistency constraint}\label{sec:appendix_samples_summarization}

See Tables \ref{tab:sum_0_samples}-\ref{tab:sum_4_samples} for summaries of 5 randomly chosen documents from our evaluation set $C_\text{test}$ generated by models fine-tuned using Conditional DPG, 3 baselines and $a$. Summaries were generated with beam search.

\subsection{Code generation with compilability constraint}\label{sec:appendix_samples_code_comp}

See Tables \ref{tab:code_0_samples}-\ref{tab:code_2_samples} for functions obtained by sampling conditioned on 3 randomly chosen signatures from our evaluation set $C_\text{test}$ generated by models fine-tuned using Conditional DPG, 3 baselines and $a$. We used pure ancestral sampling.

\subsection{Code generation with PEP8 constraint}\label{sec:appendix_samples_code_pep8}

See Tables \ref{tab:code_pep8_0_samples}-\ref{tab:code_pep8_2_samples} for functions obtained by sampling conditioned on 3 randomly chosen signatures from our evaluation set $C_\text{test}$ generated by models fine-tuned using Conditional DPG, 3 baselines and $a$. We used pure ancestral sampling.

\newpage

\begin{table*}[t]
\tiny
\begin{tabular}{p{0.5cm}p{12.8cm}}

 \multicolumn{2}{c}{{\textbf{{Source sentence $c$}}}} \\  
& ) Mr President, I would like to answer Manuel Medina Ortega' s and Efstratios Korakas' questions simultaneously, as both questions concern the risks connected with the construction of nuclear power plants, in this case, one planned in Morocco and one in Turkey. \\ \\ \\
\toprule
\textbf{$b(x,c)$} & \textbf{$x$} \\ 
\midrule

 \multicolumn{2}{c}{\textbf{Translation generated by the original model $a$}} \\  
0 & 
) Monsieur le Président, je voudrais répondre simultanément aux questions de Manuel Medina Ortega et d' Efstratios Korakas, étant donné que les deux questions concernent les risques liés à la construction de centrales nucléaires, en l'espèce, d’une centrale prévue au Maroc en Turquie.\\\\
 \multicolumn{2}{c}{\textbf{Translation generated by a model fine-tuned using Conditional DPG}} \\  
1 & 
) Monsieur le Président, je voudrais répondre simultanément aux questions de Manuel Medina Ortega et d' Efstratios Korakas, étant donné que les deux questions concernent les risques liés à la construction de centrales nucléaires, en l'espèce, 1 planifiée au Maroc, 1 en Turquie.\\\\
 \multicolumn{2}{c}{\textbf{Translation generated by a model fine-tuned using DPG}} \\  
0 & 
) Monsieur le Président, je voudrais répondre simultanément aux questions de Manuel Medina Ortega et d' Efstratios Korakas, car les deux questions concernent les risques liés à la construction de centrales nucléaires, dans ce cas-ci, l'une planifiée au Maroc, la seconde en Turquie.\\\\
 \multicolumn{2}{c}{\textbf{Translation generated by a model fine-tuned using Ziegler}} \\  
0 & 
) Monsieur le Président, je voudrais répondre simultanément aux questions de Manuel Medina Ortega et d' Efstratios Korakas, étant donné que les deux questions concernent les risques liés à la construction de centrales nucléaires, en l'espèce, d’une centrale prévue au Maroc à l’ heure actuelle, tandis qu’ une centrale en Turquie.\\\\
 \multicolumn{2}{c}{\textbf{Translation generated by a model fine-tuned using Reinforce}} \\  
1 & 
) Monsieur Président Président, Monsieur le Président, d?,, Monsieur Président, Herr Präsident, ich m 2 2, 1, 3 3, 2 2 1 1 1 2 1 2 2 2 3 1 1 3 2 2 4 1 1 5 1 1 7 1 1 4 2 2 7 7 7 1 2 3 3 2 1 3 1 2 4 2 1 7 7 8 8 1 1 8 8 8 7 7 2 2 5 2 2 8 8 2 1 4 1 2 7, 4 4 4 2\\\\
\bottomrule
\end{tabular}
\caption{\small{Translations generated by beam search on $\pit(\cdot|c)$: models fine-tuned to satisfy a terminology consistency constraints (translating numeral nouns as digits).}\label{tab:trans_0_samples}}
\end{table*}

\begin{table*}[t]
\tiny
\begin{tabular}{p{0.5cm}p{12.8cm}}

 \multicolumn{2}{c}{{\textbf{{Source sentence $c$}}}} \\  
& - (DE) The position of the non-attached Members on the content of the first two decisions of Parliament regarding the Statute for Members has not changed. \\ \\ \\
\toprule
\textbf{$b(x,c)$} & \textbf{$x$} \\ 
\midrule

 \multicolumn{2}{c}{\textbf{Translation generated by the original model $a$}} \\  
0 & 
- (DE) La position des députés non affiliés sur le contenu des deux premières décisions du Parlement concernant le statut de députés n'a pas changé.\\\\
 \multicolumn{2}{c}{\textbf{Translation generated by a model fine-tuned using Conditional DPG}} \\  
1 & 
- (DE) La position des députés non inscrits sur le contenu des 2 premières décisions du Parlement concernant le statut de députés n'a pas changé.\\\\
 \multicolumn{2}{c}{\textbf{Translation generated by a model fine-tuned using DPG}} \\  
0 & 
- (DE) La position des députés non inscrits sur le contenu des deux premières décisions du Parlement concernant le statut de députés n'a pas changé.\\\\
 \multicolumn{2}{c}{\textbf{Translation generated by a model fine-tuned using Ziegler}} \\  
0 & 
- (DE) La position des députés non affiliés sur le contenu des deux premières décisions du Parlement concernant le statut de députés n'a pas changé.\\\\
 \multicolumn{2}{c}{\textbf{Translation generated by a model fine-tuned using Reinforce}} \\  
1 & 
- (DE) La position des 2 2 2 - 2, (DE ),, 2 2 3 2 2 1 2 2 5 5 5 2 :, 1 1 1 2., 5 5 1 1, 6 2 2 4 2 2 6 5 5, 3 2 5 2 2 7 5 5 3 2 1 : 1 2 3 3 3 2 3, 4 5 5 6 6 6 5 2 5 : 2 5 1 2 1 1 5 2 3 5 5 4 2 5 6\\\\
\bottomrule
\end{tabular}
\caption{\small{Translations generated by beam search on $\pit(\cdot|c)$: models fine-tuned to satisfy a terminology consistency constraints (translating numeral nouns as digits).}\label{tab:trans_1_samples}}
\end{table*}

\begin{table*}[t]
\tiny
\begin{tabular}{p{0.5cm}p{12.8cm}}

 \multicolumn{2}{c}{{\textbf{{Source sentence $c$}}}} \\  
& - (FR) I have personally listened to all the speeches by the President, Romano Prodi, and I have participated in all six hearings of the candidate Commissioners. \\ \\ \\
\toprule
\textbf{$b(x,c)$} & \textbf{$x$} \\ 
\midrule

 \multicolumn{2}{c}{\textbf{Translation generated by the original model $a$}} \\  
0 & 
J'ai personnellement écouté toutes les interventions du président, Romano Prodi, et j'y ai participé aux six auditions des candidats aux commissaires.\\\\
 \multicolumn{2}{c}{\textbf{Translation generated by a model fine-tuned using Conditional DPG}} \\  
1 & 
J'ai personnellement écouté toutes les interventions du président, Romano Prodi, et j'y ai participé aux 6 auditions des candidats aux commissaires.\\\\
 \multicolumn{2}{c}{\textbf{Translation generated by a model fine-tuned using DPG}} \\  
0 & 
J'ai personnellement écouté toutes les interventions du président, Romano Prodi, et j'y ai participé aux six auditions des commissaires candidats.\\\\
 \multicolumn{2}{c}{\textbf{Translation generated by a model fine-tuned using Ziegler}} \\  
0 & 
J'ai personnellement écouté toutes les interventions du président, Romano Prodi, et j'y ai participé aux six auditions des candidats aux commissaires.\\\\
 \multicolumn{2}{c}{\textbf{Translation generated by a model fine-tuned using Reinforce}} \\  
1 & 
- FR) J' m 2 2 2, 2 2 3 3 3, 1,, 5 5 5 :, 3 3 5 5 1 1 1 2 2 1 2 1 1 3 3 2 2 5 5 3 3 1 1 5 5 2 1 3 2 3 1 2 3 2 1 5 2 2 4 3 3 4 5 5 7 7 7 5 5 6 6 6 5 5 4 2 2 7 7 6 6 7 7, 6 6 3 3 7 7 3 5 3 5 7 5 7 6 5\\\\
\bottomrule
\end{tabular}
\caption{\small{Translations generated by beam search on $\pit(\cdot|c)$: models fine-tuned to satisfy a terminology consistency constraints (translating numeral nouns as digits).}\label{tab:trans_2_samples}}
\end{table*}

\begin{table*}[t]
\tiny
\begin{tabular}{p{0.5cm}p{12.8cm}}

 \multicolumn{2}{c}{{\textbf{{Source sentence $c$}}}} \\  
& A decision will probably be taken to extend the existing measures against Burma, including visa restrictions, for a further six months. \\ \\ \\
\toprule
\textbf{$b(x,c)$} & \textbf{$x$} \\ 
\midrule

 \multicolumn{2}{c}{\textbf{Translation generated by the original model $a$}} \\  
0 & 
Une décision sera probablement prise pour prolonger de six mois les mesures en vigueur contre la Birmanie, y compris les restrictions à la délivrance de visas.\\\\
 \multicolumn{2}{c}{\textbf{Translation generated by a model fine-tuned using Conditional DPG}} \\  
1 & 
Une décision sera probablement prise pour prolonger de 6 mois les mesures en vigueur contre la Birmanie, y compris les restrictions à la délivrance de visas.\\\\
 \multicolumn{2}{c}{\textbf{Translation generated by a model fine-tuned using DPG}} \\  
1 & 
On décidera probablement de proroger de 6 mois les mesures en vigueur contre la Birmanie, y compris les restrictions à la délivrance de visas.\\\\
 \multicolumn{2}{c}{\textbf{Translation generated by a model fine-tuned using Ziegler}} \\  
0 & 
On décidera probablement de proroger de six mois les mesures en vigueur contre la Birmanie, y compris les restrictions à la délivrance de visas.\\\\
 \multicolumn{2}{c}{\textbf{Translation generated by a model fine-tuned using Reinforce}} \\  
1 & 
Il y a 6 6 6 7 7 7, 6 6 2 6 6, 2 2 2 1 1 1 2 2, y 2 1 2 1, 1 1 - 1 2 -, 4 2 2 3, 5 2 2 4, 7 7 2 2 5, 3 2 2 7 - 2 2 8, 8 - 5 - 7'' 2'1 1 3 - 4 - 3'5'7 7 8 7 7 5 \\\\
\bottomrule
\end{tabular}
\caption{\small{Translations generated by beam search on $\pit(\cdot|c)$: models fine-tuned to satisfy a terminology consistency constraints (translating numeral nouns as digits).}\label{tab:trans_3_samples}}
\end{table*}

\begin{table*}[t]
\tiny
\begin{tabular}{p{0.5cm}p{12.8cm}}

 \multicolumn{2}{c}{{\textbf{{Source sentence $c$}}}} \\  
& A particularly thorny question, and one that has been cause for most debate beforehand, is that of Turkey. \\ \\ \\
\toprule
\textbf{$b(x,c)$} & \textbf{$x$} \\ 
\midrule

 \multicolumn{2}{c}{\textbf{Translation generated by the original model $a$}} \\  
0 & 
Une question particulièrement épineuse, qui a suscité la plupart des débats préalables, est celle de la Turquie.\\\\
 \multicolumn{2}{c}{\textbf{Translation generated by a model fine-tuned using Conditional DPG}} \\  
0 & 
Une question particulièrement épineuse, qui a suscité la plupart des débats antérieurs, est celle de la Turquie.\\\\
 \multicolumn{2}{c}{\textbf{Translation generated by a model fine-tuned using DPG}} \\  
0 & 
Une question particulièrement épineuse, qui a suscité la plupart des débats préalables, est celle de la Turquie.\\\\
 \multicolumn{2}{c}{\textbf{Translation generated by a model fine-tuned using Ziegler}} \\  
0 & 
Une question particulièrement épineuse, qui a suscité la plupart des débats préalables, est celle de la Turquie.\\\\
 \multicolumn{2}{c}{\textbf{Translation generated by a model fine-tuned using Reinforce}} \\  
1 & 
Une question particulièrement épineuse, et 1, 1 1 1 2 2 2 1 2 1 1 3 2 2 3 2 1 3 1 1 5 1 1 4 2 2 5 1 2 3 3 2 3 1 2 5 2 2 4 2 1 5 5 1 3 3 3 1 3 5 5 5 2 1 4 1 2 4 1 1 : 1 1 6 1 2, 2 2 : 2 3 5 1 5 2 5 5 3 2 5 3 3 5 2 3 4 5 5, 5 5 4 2 5 4\\\\
\bottomrule
\end{tabular}
\caption{\small{Translations generated by beam search on $\pit(\cdot|c)$: models fine-tuned to satisfy a terminology consistency constraints (translating numeral nouns as digits).}\label{tab:trans_4_samples}}
\end{table*}

\begin{table*}[t]
\tiny
\begin{center}
\begin{tabular}{p{0.5cm}p{12.8cm}}

 \multicolumn{2}{c}{{\textbf{{Source document $c$}}}} \\  
& (CNN)"Success Kid" is likely the Internet's most famous baby. You've seen him in dozens of memes, fist clenched in a determined look of persevering despite the odds. Success Kid -- now an 8-year-old named Sammy Griner -- needs a little bit of that mojo to rub off on his family.  His dad, Justin, needs a kidney transplant. About a week ago, Laney Griner, Justin's wife and Sammy's mother, created a GoFundMe campaign with a goal of \$75,000 to help cover the medical expenses that go along with a kidney transplant. The campaign is already a success. By Wednesday it had topped its goal. Griner told The Daily Dot that her husband was diagnosed with kidney disease in 2006 and suffered complete kidney failure three years later. "One can only survive with no natural kidney function ... for so long," Laney Griner said. "His energy and mood are affected; he can no longer work, and he spends 12 hours a week in dialysis clinic. "Having been on dialysis for this long greatly increases his risks of developing further complications. The only way to save his life is to get a transplant. There's no other way around that," she said. The family doesn't know when a kidney might become available.  Their GoFundMe page has a link for potential donors. Sammy's Internet fame began in 2007 when his mom posted a picture of him on a beach with a fist full of sand and a satisfied look on his face.  Myspace picked it up, so did Reddit. The rest is Internet history.  Success just seems to run in some families. \\ \\ \\
\toprule
\textbf{$b(x,c)$} & \textbf{$x$} \\ 
\midrule

 \multicolumn{2}{c}{\textbf{Summary generated by the original model $a$}} \\  
0 & 
"Success Kid" is the internet's most famous baby. his dad, \textcolor{red}{Justin}, needs a kidney transplant. a week ago, his wife and mother created a \textcolor{red}{goFundMe} campaign.\\\\
 \multicolumn{2}{c}{\textbf{Summary generated by a model fine-tuned using Conditional DPG}} \\  
0 & 
"Success Kid" is the internet's most famous baby. his dad, \textcolor{red}{Justin}, needs a kidney transplant. about a week ago, \textcolor{red}{Laney Griner}, his wife and mother created a \textcolor{red}{goFundMe} campaign.\\\\
 \multicolumn{2}{c}{\textbf{Summary generated by a model fine-tuned using DPG}} \\  
0 & 
"Success Kid" is the internet's most famous baby. his mom posted a picture of him on a beach with a fist full of sand and a satisfied look on his face. a \textcolor{red}{goFundMe} campaign has a goal of \$75,000 to help cover medical expenses.\\\\
 \multicolumn{2}{c}{\textbf{Summary generated by a model fine-tuned using Ziegler}} \\  
0 & 
"Success Kid" is the internet's most famous baby. his dad, \textcolor{red}{Justin}, needs a kidney transplant. a week ago, his wife and mother created a \textcolor{red}{goFundMe} campaign.\\\\
 \multicolumn{2}{c}{\textbf{Summary generated by a model fine-tuned using Reinforce}} \\  
1 & 
Success Kid -- now an 8-year-old named \textcolor{red}{Sammy Griner} -- needs a little bit of that mojo to rub off on his family. \textcolor{red}{Laney Griner}, textcolor{red}{Justin}'s wife and \textcolor{red}{Sammy}'s mother, created a \textcolor{red}{GoFundMe} campaign with a goal of \$75,000. \textcolor{red}{Griner} told \textcolor{red}{The Daily Dot} that her husband was diagnosed with kidney disease in 2006 and suffered complete kidney failure three years later. The family doesn't know when a kidney might become available. \textcolor{red}{Sammy}’s\\\\
\bottomrule
\end{tabular}
\end{center}
\caption{\small{Summaries generated by beam search on $\pit(\cdot|c)$: models fine-tuned to satisfy entity-level factual consistency constraint. Named entities in summaries are highlighted in \textcolor{red}{red}. }\label{tab:sum_0_samples}}
\end{table*}

\begin{table*}[t]
\tiny
\begin{center}
\begin{tabular}{p{0.5cm}p{12.8cm}}
    
 \multicolumn{2}{c}{{\textbf{{Source document $c$}}}} \\  
& A picture, believed to be the only image of the Civil War ironclad, the CSS Georgia has been revealed to be a fake created in a teenage hoax using a 2ft model. John Potter, from Savannah, has admitted forging the picture with his brother in the 1980s and placing it in a frame which now holds a picture of his dead dog. He then passed on the image on to the Georgia Historical Society and the photo became an unofficial part of the ship's history even though it was never authenticated. Scroll down for video . John Potter (pictured), from Savannah, has admitted forging the picture with his brother in the 1980s and placing the picture in a frame which now holds a picture of his dead dog . In 1986 he fibbed that he was at a yard sale when he found the photograph in an antique frame.Inscribed on the back of the frame, he claimed, was 'CSS Georgia.' He told historians that he didn't have the \$175 the owner wanted . The picture Mr Potter handed over to the society was actually a picture of a picture. In 1986 he fibbed that he was at a yard sale when he found the photograph in an antique frame. Inscribed on the back of the frame, he claimed, was 'CSS Georgia.' He told historians that he didn't have the \$175 the owner wanted, so he took a photo of it and then mailed it to historical groups in Savannah. Potter has now admitted the sham and explained how he falsified the image. When he was a teenager in Savannah, Potter, his brother Jeffrey and a friend shot a short 8mm movie about the building — and destruction — of the CSS Georgia in a fictional battle with Union troops. For the movie, they built an 18-foot long boat of plywood and Styrofoam, as well as a smaller 2-foot model. They based the design, in part, on his grandfather's recollections of details passed down by word of mouth through generations of their family. Potter also used an illustration of the ironclad he found on a postcard. To create the fake image Potter's younger brother put on a coat and straw hat went out to a marsh with a cane fishing pole and Potter took a photo. He took another photo of the 2-foot model and cut out the boat's image, glued it onto the photo of his brother, then used dirt and glue to create the illusion of a photo faded by age and stained by water or chemicals. Evidence of the hoax: All elements of the fakery were snapped by Potter at one time. The small model boat here appears alongside the false picture, polaroids of it and a 1984 copy of Mad Magazine . He bought an old picture frame and beat it up even further. He put the photo in it. Then he drove 120 miles to a yard sale in Waycross, Georgia, put the picture down and took a Polaroid of it. He laughs now, when he remembers that it had seemed so important that he actually do this at a yard sale, so at least that part would be true. 'Who knows what goes through the mind of a kid,' he said. The US Army Corps of Engineers appealed for information about the picture as it embarked on a project to raise the ironclad . Potter sent out the photo to historical groups, setting off the sporadic, and fruitless, search for a CSS Georgia photo that he now says never existed. As the Army Corps of Engineers embarked this year on a \$14 million project to raise the wreckage of the real CSS Georgia from the river, archaeologists publicised the image online and in news stories hoping to track down the original photo. 'Wanted: A Photo Of This Confederate Battleship,' read the headline above the image on the Army Corps website. 'There are a lot of characteristics in the photograph that lead us to believe it is the CSS Georgia,' Julie Morgan, the Army Corps archaeologist in charge of raising the ironclad's remains, said in a February interview. 'On the other side, there are some skeptics who believe it's a complete fake.' Robert Holcombe, former curator of the National Civil War Naval Museum in Columbus, Georgia, said in February that while the original photograph would be needed to confirm if the image was authentic, he believed it was real. 'Most people seem to think so,' he said. 'Or else it's an awfully good fake'. The peeling gilt frame that once held the disputed photo, is now filled with a portrait of Potter's deceased pug, Puggy Van Dug. He never became a successful special effects artist with the exception of the one faked photo. The peeling gilt frame that once held the disputed photo, is now filled with a portrait of Potter's deceased pug, Puggy Van Dug . \\ \\ \\
\toprule
\textbf{$b(x,c)$} & \textbf{$x$} \\ 
\midrule

 \multicolumn{2}{c}{\textbf{Summary generated by the original model $a$}} \\  
0 & 
\textcolor{red}{John Potter}, from \textcolor{red}{Savannah}, has admitted forging the picture with his brother in the 1980s and placing it in a frame which now holds a picture of his dead dog. he then passed on the image on to the \textcolor{red}{Georgia Historical Society} and the photo became an unofficial part of the ship's history even though it was never authenticated. in 1986 he fibbed that he was at a yard sale when he found the photograph in an antique frame.\\\\
 \multicolumn{2}{c}{\textbf{Summary generated by a model fine-tuned using Conditional DPG}} \\  
0 & 
\textcolor{red}{John Potter}, from \textcolor{red}{Savannah}, has admitted forging the picture with his brother in the 1980s and placing it in a frame which now holds a picture of his dead dog. he passed on the image on to the \textcolor{red}{Georgia Historical Society} and the photo became an unofficial part of the ship's history even though it was never authenticated. in 1986 he fibbed that he was at a yard sale when he found the photograph in an antique frame.inscribed on the back of the frame, he claimed,\\\\
 \multicolumn{2}{c}{\textbf{Summary generated by a model fine-tuned using DPG}} \\  
0 & 
\textcolor{red}{John Potter}, from \textcolor{red}{Savannah}, has admitted forging the picture with his brother in the 1980s and placing it in a frame which now holds a picture of his dead dog. the picture became an unofficial part of the ship's history even though it was never authenticated. in 1986 he fibbed that he was at a yard sale when he found the photograph in an antique frame.inscribed on the back of the frame, he claimed, was \textcolor{red}{'CSS Georgia'}\\\\
 \multicolumn{2}{c}{\textbf{Summary generated by a model fine-tuned using Ziegler}} \\  
0 & 
\textcolor{red}{John Potter}, from \textcolor{red}{Savannah}, has admitted forging the picture with his brother in the 1980s and placing it in a frame which now holds a picture of his dead dog. he then passed on the image on to the \textcolor{red}{Georgia Historical Society} and the photo became an unofficial part of the ship's history even though it was never authenticated. in 1986 he fibbed that he was at a yard sale when he found the photograph in an antique frame.\\\\
 \multicolumn{2}{c}{\textbf{Summary generated by a model fine-tuned using Reinforce}} \\  
0 & 
\textcolor{red}{John Potter}, from \textcolor{red}{Savannah}, has admitted forging the picture with his brother in the 1980s and placing it in a frame which now holds a picture of his dead dog. He told historians that \textcolor{red}{Mr Potter} handed over to the \textcolor{red}{Georgia Historical Society} and the photo became an unofficial part of the ship's history. \textcolor{red}{Potter} has now admitted the sham and explained how he falsified the image when he was a teenager in \textcolor{red}{Savannah}, \textcolor{red}{Potter}, his brother \textcolor{red}{Jeffrey} and a friend shot a short 8mm movie about\\\\
\bottomrule
\end{tabular}
\end{center}
\caption{\small{Summaries generated by beam search on $\pit(\cdot|c)$: models fine-tuned to satisfy entity-level factual consistency constraint. Named entities in summaries are highlighted in \textcolor{red}{red}.}\label{tab:sum_1_samples}}
\end{table*}

\begin{table*}[t]
\tiny
\begin{center}
\begin{tabular}{p{0.5cm}p{12.8cm}}

 \multicolumn{2}{c}{{\textbf{{Source document $c$}}}} \\  
& Everton's Steven Pienaar has admitted he considered retirement as frustration took its toll during his injury-ravaged season. The influential 33-year-old midfielder Pienaar has been dogged by groin and knee injuries this season limiting him to just 11 appearances. He returned to action from his latest setback in the 1-1 draw with Swansea City earlier this month but muscle fatigue ruled him out of the win over Burnley last weekend. Everton midfielder Steven Pienaar is held back by Swansea's Ki Sung-Yueng at the Liberty Stadium . Pienaar said: 'At one stage, I thought I had better just hang my boots up and call it a day but on the other side I was just thinking that I enjoy going in and seeing the guys so I just had to stay strong and that kept me going. 'When you are at home not coming in for training you feel very down but as soon as I walk through the door, there’s always the camaraderie in the group, there is always fun. 'Even if you are injured, you can always laugh and it keeps you going. Just to be among the players, it’s kept me going.' Pienaar has made just 11 appearances for the Toffees this season due to groin and knee injuries . \\ \\ \\
\toprule
\textbf{$b(x,c)$} & \textbf{$x$} \\ 
\midrule

 \multicolumn{2}{c}{\textbf{Summary generated by the original model $a$}} \\  
0 & 
the 33-year-old midfielder has been dogged by groin and knee injuries. he has made just 11 appearances for the toffees this season. \textcolor{red}{pienaar} returned to action from his latest setback in the 1-1 draw.\\\\
 \multicolumn{2}{c}{\textbf{Summary generated by a model fine-tuned using Conditional DPG}} \\  
0 & 
\textcolor{red}{Steven Pienaar} has been dogged by groin and knee injuries this season. the 33-year-old midfielder has made just 11 appearances for the toffees. he returned to action from his latest setback in the 1-1 draw with \textcolor{red}{Swansea city} earlier this month.\\\\
 \multicolumn{2}{c}{\textbf{Summary generated by a model fine-tuned using DPG}} \\  
0 & 
the 33-year-old midfielder has been dogged by groin and knee injuries this season. he returned to action from his latest setback in the 1-1 draw with \textcolor{red}{Swansea city}. \textcolor{red}{pienaar} has made just 11 appearances for the toffees this season due to injury.\\\\
 \multicolumn{2}{c}{\textbf{Summary generated by a model fine-tuned using Ziegler}} \\  
0 & 
the 33-year-old midfielder has been dogged by groin and knee injuries. he has made just 11 appearances for the toffees this season. \textcolor{red}{pienaar} returned to action from his latest setback in the 1-1 draw.\\\\
 \multicolumn{2}{c}{\textbf{Summary generated by a model fine-tuned using Reinforce}} \\  
0 & 
\textcolor{red}{Everton}'s \textcolor{red}{Steven Pienaar} has admitted he considered retirement as frustration took its toll during his injury-ravaged season. The influential 33-year-old midfielder \textcolor{red}{Pienanar} has been dogged by groin and knee injuries this season for the \textcolor{red}{Toffees}. He returned to action from his latest setback in a 1-1 draw with \textcolor{red}{Swansea City} earlier this month but muscle fatigue ruled him out of the win over \textcolor{red}{Burnley} last weekend. \textcolor{red}{Pienar} said Pien\\\\
\bottomrule
\end{tabular}
\end{center}
\caption{\small{Summaries generated by beam search on $\pit(\cdot|c)$: models fine-tuned to satisfy entity-level factual consistency. Named entities in summaries are highlighted in \textcolor{red}{red}. constraint}\label{tab:sum_2_samples}}
\end{table*}

\begin{table*}[t]
\tiny
\begin{center}
\begin{tabular}{p{0.5cm}p{12.8cm}}

 \multicolumn{2}{c}{{\textbf{{Source document $c$}}}} \\  
& Expat: Detectives are investigating the killing of David King (pictured), a 70-year-old pensioner from Newham, east London, who retired to Normandy . A fight over stolen vegetables may have led to a British expat being murdered and dumped at the bottom of a well, French police fear. The macabre theory has been outlined by detectives investigating the killing of David King, a 70-year-old pensioner from Newham, east London, who retired to Normandy 15 years ago. His body was found by sniffer dogs last week in the picturesque hamlet of Pierres, south-west of Caen, and an unnamed 28-year-old Frenchman has been charged with his murder. The alleged killer was living rough in the area, and is thought to have been behind a number of thefts in the area. Now prosecutor Carole Etienne has said that Mr King, a keen gardener, may well have confronted the man over stolen vegetables. Ms Etienne, who is leading the police enquiry, said: ‘Among the vague attempts at an explanation given by the accused, we are looking at the possibility of a fight over the theft of foodstuffs. ‘In one night, a whole plot of leeks might disappear. One person saw four rabbits vanishing overnight.’ Ms Etienne added: ‘People were getting more and more angry, and some were even threatening to defend their plots with shotguns’. Neighbours of Mr King, who was hugely proud of his vegetable garden, told Le Parisien newspaper that a description of the thief corresponds to the alleged murderer. Discovery: In February, Mr King's car, a Renault Scenic (above) was found parked in Vire, a nearby town, but there was no sign of any body . Family and neighbours of Mr King have attacked detectives for allowing his suspected murderer to remain at large for six months. Interpol, the international police organisation, had initially refused to open a missing person’s enquiry. Instead they believed Mr King had travelled to Australia to see his daughter, Sandie Ray. Mr King had been living in France for 15 years . Ms Ray, who lives in Perth, Australia, said her father’s passport details had been mixed up with another David King, who had indeed travelled to Australia from France. She said this was known by November last year, but ‘the French authorities still hadn’t been formally notified of this via Interpol until approximately three months later’. Ms Ray said a quicker enquiry would have avoided ‘lots of anxiety and frustration for our family and dads friends.’ John King, Mr King’s son, who lives in Brighton, East Sussex, said the botched investigation had been a ‘bureaucratic nightmare’ for all concerned. Other expats living in the area said it was ‘hugely frightening’ to have a suspected murderer living in their midst while the operation went on. ‘This is an isolated part of the world, and everyone is potentially vulnerable to attack,’ said one. ‘Detectives should have worked out what was going on far quicker. The slow speed of the enquiry was unacceptable.' In February, Mr King's car, a Renault Scenic was found parked in Vire, a nearby town, but there was no sign of any body. \\ \\ \\
\toprule
\textbf{$b(x,c)$} & \textbf{$x$} \\ 
\midrule

 \multicolumn{2}{c}{\textbf{Summary generated by the original model $a$}} \\  
0 & 
the macabre theory has been outlined by detectives investigating the killing of a 70-year-old pensioner from \textcolor{red}{newham, east London}. his body was found by sniffer dogs last week in the picturesque \textcolor{red}{hamlet of Pierres}, south-west of \textcolor{red}{Caen}. an unnamed 28-year old frenchman has been charged with his murder.\\\\
 \multicolumn{2}{c}{\textbf{Summary generated by a model fine-tuned using Conditional DPG}} \\  
1 & 
the macabre theory has been outlined by detectives investigating the killing of \textcolor{red}{David King}, 70, from \textcolor{red}{newham}, east \textcolor{red}{London}, who retired to \textcolor{red}{Normandy} 15 years ago. his body was found by sniffer dogs last week in the picturesque hamlet of \textcolor{red}{Pierres}, south-west of \textcolor{red}{Caen}. an unnamed 28-year-old frenchman has been charged with his murder. prosecutor \textcolor{red}{Carole Etienne} has said that \textcolor{red}{Mr King}, a keen gardener, may well have confronted the man over over stolen vegetables\\\\
 \multicolumn{2}{c}{\textbf{Summary generated by a model fine-tuned using DPG}} \\  
0 & 
the macabre theory has been outlined by detectives investigating the killing of a 70-year-old pensioner from \textcolor{red}{newham}, east \textcolor{red}{London}, who retired to Normandy 15 years ago. his body was found by sniffer dogs last week in the picturesque hamlet of \textcolor{red}{Pierres}, south-west of \textcolor{red}{caen}, and an unnamed 28-year old frenchman has been charged with his murder. prosecutor \textcolor{red}{Carole Etienne} has said that the alleged killer may well have confronted the man over stolen vegetables\\\\
 \multicolumn{2}{c}{\textbf{Summary generated by a model fine-tuned using Ziegler}} \\  
0 & 
the macabre theory has been outlined by detectives investigating the killing of a 70-year-old pensioner from \textcolor{red}{newham}, east \textcolor{red}{London}. his body was found by sniffer dogs last week in the picturesque hamlet of \textcolor{red}{Pierres}, south-west of \textcolor{red}{Caen}. an unnamed 28-year old frenchman has been charged with his murder.\\\\
 \multicolumn{2}{c}{\textbf{Summary generated by a model fine-tuned using Reinforce}} \\  
1 & 
\textcolor{red}{David King}, a 70-year-old pensioner from \textcolor{red}{Newham}, east \textcolor{red}{London}, who retired to \textcolor{red}{Normandy} 15 years ago. His body was found by sniffer dogs last week in the picturesque hamlet of \textcolor{red}{Pierres}, south-west of \textcolor{red}{Caen}, and an unnamed Frenchman has been charged with his murder. prosecutor \textcolor{red}{Carole Etienne} has said \textcolor{red}{Mr King}, keen gardener, may well have confronted the man over stolen vegetables, Neighbours of \textcolor{red}{Mr King}. \textcolor{red}{Interpol}, international police organisation\\\\
\bottomrule
\end{tabular}
\end{center}
\caption{\small{Summaries generated by beam search on $\pit(\cdot|c)$: models fine-tuned to satisfy entity-level factual consistency constraint. Named entities in summaries are highlighted in \textcolor{red}{red}. }\label{tab:sum_3_samples}}
\end{table*}

\begin{table*}[t]
\tiny
\begin{center}
\begin{tabular}{p{0.5cm}p{12.8cm}}

 \multicolumn{2}{c}{{\textbf{{Source document $c$}}}} \\  
& German legend Franz Beckenbauer has accused Bayern Munich stars of playing as if they had taken 'sleeping pills' in their midweek defeat by Porto. The Bundesliga champions conceded twice in the opening 10minutes before losing 3-1 to the Portuguese in the first leg of their Champions League quarter-final on Wednesday. Der Kaiser may be a brand ambassador for the club but he couldn't hide his feelings after the game, when he criticised defender Dante for playing as if he were wearing 'ski boots' before turning on the entire team. Franz Beckenbauer was speaking in New York as part of Bayern Munich's media agreement with MSN . New York Cosmos legends Beckenbauer and Pele pose together after launching the team's spring season . Ricardo Quaresma scored the opening goal from the spot in Porto's 3-1 defeat of the Germans . Beckenbauer accused Brazilian defender Dante (left) of playing as if he were wearing 'ski boots' Speaking to reporters in New York to mark the club's new media agreement with MSN, Beckenbauer said: 'It was one of those days, all the players didn't show their real performance. 'After 10 minutes you are 2-0 down in the quarter-final in the Champions League, so many mistakes. I never saw this before. I thought they took sleeping pills! Porto were much faster, real power. It made me angry. Quaresma celebrates with Porto team-mates after netting his second goal inside the first 10minutes . Bayern Munich head coach Pep Guardiola goes crazy at his team's performance . Bayern players trudge off after defeat knowing they will have to vastly improve to reach the semi-finals . 'If they played like that again, no chance [they can go through]. But they can win 2-0 or go to extra-time. They have a chance. Last year we had the same problem, playing excellent in the Bundesliga and then we had two bad days and were out of the competition.' The 69-year-old did attempt to play down his comments over Dante's performance, he said: 'At half-time, I have to give a comment and that was my reaction. And I'm sorry to give him the ski boots... but he's Brazilian, and Brazilian to me means for me, technique and Pele and... Brazil. But not this.' 'If you want to win the Champions League you need a strong league with a lot of players. Internationally, Bayern Munich belongs to the top teams but to win the Champions League… you saw the game probably two days ago, we are far, far away. With this performance, you don't win anything,' he added. \\ \\ \\
\toprule
\textbf{$b(x,c)$} & \textbf{$x$} \\ 
\midrule

 \multicolumn{2}{c}{\textbf{Summary generated by the original model $a$}} \\  
0 & 
german legend claims players played as if they had taken'sleeping pills' \textcolor{red}{porto} lost 3-1 to \textcolor{red}{porto} in the first leg of their champions league quarter-final. the 69-year-old was speaking to reporters in \textcolor{red}{new york} as part of the club's media agreement with \textcolor{red}{MSN}.\\\\
 \multicolumn{2}{c}{\textbf{Summary generated by a model fine-tuned using Conditional DPG}} \\  
1 & 
german legend \textcolor{red}{Franz Beckenbauer} was speaking in \textcolor{red}{new york} as part of the club's media agreement with \textcolor{red}{MSN}. he accused defender \textcolor{red}{Dante} of playing as if he were wearing'ski boots' the 69-year-old said: 'if they played like that again, no chance [they can go through] but they can win 2-0 or go to extra-time'\\\\
 \multicolumn{2}{c}{\textbf{Summary generated by a model fine-tuned using DPG}} \\  
0 & 
the germans conceded twice in the opening 10 minutes before losing 3-1 to \textcolor{red}{porto}. the 69-year-old criticised defender defender \textcolor{red}{Dante} for playing as if he were wearing'ski boots' he said: 'if they played like that again, no chance [they can go through]'\\\\
 \multicolumn{2}{c}{\textbf{Summary generated by a model fine-tuned using Ziegler}} \\  
0 & 
german legend claims players played as if they had taken'sleeping pills' \textcolor{red}{porto} lost 3-1 to porto in the first leg of their champions league quarter-final. the 69-year-old was speaking to reporters in \textcolor{red}{new york} as part of the club's media agreement with MSN.\\\\
 \multicolumn{2}{c}{\textbf{Summary generated by a model fine-tuned using Reinforce}} \\  
1 & 
\textcolor{red}{Franz Beckenbauer} has accused \textcolor{red}{Bayern Munich} stars of playing as if they had taken'sleeping pills' the \textcolor{red}{Bundesliga} champions concede twice in opening 10minutes before losing 3-1 to the Portuguese in their Champions League quarter-final on Wednesday. \textcolor{red}{Der Kaiser} may be a brand ambassador for the club but he couldn't hide his feelings after the game. \textcolor{red}{Ricardo Quaresma} scored his opening goal from spot in \textcolor{red}{Porto}'s 3-1 defeat of the Germans. \textcolor{red}{Bayern Munich} head coach Pe\\\\
\bottomrule
\end{tabular}
\end{center}
\caption{\small{Summaries generated by beam search on $\pit(\cdot|c)$: models fine-tuned to satisfy entity-level factual consistency constraint. Named entities in summaries are highlighted in \textcolor{red}{red}.}\label{tab:sum_4_samples}}
\end{table*}

\begin{table*}[h]
\tiny
\begin{tabular}{l|l}
\textbf{$b(x,c)$} & \textbf{$[x,c]$} \\ 
\midrule

 \multicolumn{2}{c}{\textbf{Sequences sampled from the original model $a$}} \\  
0 & 

\begin{minipage}{.9\textwidth}
\begin{minted}[breaklines]{python}
def __init__(self):
    self.peneturn = TitlePane(text='Screen' if len(self.peneturn) > 0 else "Find" )
    self. gentlemen = TitlePane()
    self.managers = TitlePane()
    self.title = TitleBox(title='Title' )
    self.pane = self.hides(self.menues, menues=self.menues)
    self.instructions = Label(text= 'I' )
    self.scripture =
\end{minted}
\end{minipage}
\\
\\
\hline
\\0 & 

\begin{minipage}{.9\textwidth}
\begin{minted}[breaklines]{python}
def __init__(self):
    """
\end{minted}
\end{minipage}
\\
\\
\hline
\\0 & 

\begin{minipage}{.9\textwidth}
\begin{minted}[breaklines]{python}
def __init__(self):
        self.attach_button()
        self.eager_init()
        self.x_template = elem_self.template "x_template"
\end{minted}
\end{minipage}
\\
\\
\hline
\\
 \multicolumn{2}{c}{\textbf{Sequences sampled from a model fine-tuned using Conditional DPG}} \\  
1 & 

\begin{minipage}{.9\textwidth}
\begin{minted}[breaklines]{python}
def __init__(self):
        self.context = 100
\end{minted}
\end{minipage}
\\
\\
\hline
\\1 & 

\begin{minipage}{.9\textwidth}
\begin{minted}[breaklines]{python}
def __init__(self):
        self.html_to_css_css2 = response()
        self.html_to_content_css3 = response()
        self.html_to_css3 = response()
\end{minted}
\end{minipage}
\\
\\
\hline
\\1 & 

\begin{minipage}{.9\textwidth}
\begin{minted}[breaklines]{python}
def __init__(self):
    super(DispAdmin, self).__init__(self)
\end{minted}
\end{minipage}
\\
\\
\hline
\\
 \multicolumn{2}{c}{\textbf{Sequences sampled from a model fine-tuned using DPG}} \\  
1 & 

\begin{minipage}{.9\textwidth}
\begin{minted}[breaklines]{python}
def __init__(self):
        pass
\end{minted}
\end{minipage}
\\
\\
\hline
\\1 & 

\begin{minipage}{.9\textwidth}
\begin{minted}[breaklines]{python}
def __init__(self):
    stuff = createObject()
    self.pushItem(stuff, __get_numbers, 'numbers', - '0', 'tnC' )
    stuff.name = '#%s';
    self.pop()
\end{minted}
\end{minipage}
\\
\\
\hline
\\1 & 

\begin{minipage}{.9\textwidth}
\begin{minted}[breaklines]{python}
def __init__(self):
        self.gameDataGroup = []
        self.playDataSetGroup = ['set']
        self.lpsCreate = None
        self.playerDataSetGroup = [class_DataGroup]
        if self.gameDataGroup is not None:
            self.gameDataGroup.append(self.gameGroup)
\end{minted}
\end{minipage}
\\
\\
\hline
\\
 \multicolumn{2}{c}{\textbf{Sequences sampled from a model fine-tuned using Ziegler}} \\  
1 & 

\begin{minipage}{.9\textwidth}
\begin{minted}[breaklines]{python}
def __init__(self):
        self.headers_headers_headers_headers_headers_headers_headers_headers_headers_headers_headers_
\end{minted}
\end{minipage}
\\
\\
\hline
\\1 & 

\begin{minipage}{.9\textwidth}
\begin{minted}[breaklines]{python}
def __init__(self):
    self.shape_children_children_children_children_children_children_children_children_
\end{minted}
\end{minipage}
\\
\\
\hline
\\1 & 

\begin{minipage}{.9\textwidth}
\begin{minted}[breaklines]{python}
def __init__(self):
    self._items_items_items_items_items_items_items_items_items_items_items_items_items_items_items_items_items_items_
\end{minted}
\end{minipage}
\\
\\
\hline
\\
 \multicolumn{2}{c}{\textbf{Sequences sampled from a model fine-tuned using Reinforce}} \\  
1 & 

\begin{minipage}{.9\textwidth}
\begin{minted}[breaklines]{python}
def __init__(self):
    self.screen = screen
\end{minted}
\end{minipage}
\\
\\
\hline
\\1 & 

\begin{minipage}{.9\textwidth}
\begin{minted}[breaklines]{python}
def __init__(self):
    self.value = None
    self.shadow = None
    self.showTitles()
    self.context = None
\end{minted}
\end{minipage}
\\
\\
\hline
\\1 & 

\begin{minipage}{.9\textwidth}
\begin{minted}[breaklines]{python}
def __init__(self):
    self.description = ''
    self.title    = ''
\end{minted}
\end{minipage}
\\
\\
\bottomrule
\end{tabular}
\caption{\small{Samples obtained from $\pit(\cdot|c)$ with $c =$ \texttt{def \_\_init\_\_(self)} fine-tuned to satisfy a compilability constraint}\label{tab:code_0_samples}}
\end{table*}

\begin{table*}[h]
\tiny
\begin{tabular}{l|l}
\textbf{$b(x,c)$} & \textbf{$[x,c]$} \\ 
\midrule

 \multicolumn{2}{c}{\textbf{Sequences sampled from the original model $a$}} \\  
0 & 

\begin{minipage}{.9\textwidth}
\begin{minted}[breaklines]{python}
def __init__(self,* args,** kwargs):
    self.algorithm = algorithm.__algorithm
    self.scope = scope.scope
    instance = remotely(self.altstr, 1.4)
    if (num == 3 and instance.level >= 2):
        self.alg = alg
        self.algorithm=algorithm
        self.scope = sprintf(self.scope, "%r\n", "%d\n")
        return *self.alg
\end{minted}
\end{minipage}
\\
\\
\hline
\\1 & 

\begin{minipage}{.9\textwidth}
\begin{minted}[breaklines]{python}
def __init__(self,* args,** kwargs):
        super(SceneApplication,self).__init__(args,kwargs,self.plane)
\end{minted}
\end{minipage}
\\
\\
\hline
\\0 & 

\begin{minipage}{.9\textwidth}
\begin{minted}[breaklines]{python}
def __init__(self,* args,** kwargs):
    from base import *
    from dict import ddarray
    from. import *
    from dbase import *
    from dbase.serial.calls import *
    from. import __file__ import *
    from. __del__ import *
\end{minted}
\end{minipage}
\\
\\
\hline
\\
 \multicolumn{2}{c}{\textbf{Sequences sampled from a model fine-tuned using Conditional DPG}} \\  
0 & 

\begin{minipage}{.9\textwidth}
\begin{minted}[breaklines]{python}
def __init__(self,* args,** kwargs):

\end{minted}
\end{minipage}
\\
\\
\hline
\\0 & 

\begin{minipage}{.9\textwidth}
\begin{minted}[breaklines]{python}
def __init__(self,* args,** kwargs):
    from PIL import *
\end{minted}
\end{minipage}
\\
\\
\hline
\\0 & 

\begin{minipage}{.9\textwidth}
\begin{minted}[breaklines]{python}
def __init__(self,* args,** kwargs):

\end{minted}
\end{minipage}
\\
\\
\hline
\\
 \multicolumn{2}{c}{\textbf{Sequences sampled from a model fine-tuned using DPG}} \\  
1 & 

\begin{minipage}{.9\textwidth}
\begin{minted}[breaklines]{python}
def __init__(self,* args,** kwargs):
        raise ValueError('' + self.__class__.__name__ +'does not exist'
                        + self.__class__.__name__,__LINE__)
\end{minted}
\end{minipage}
\\
\\
\hline
\\1 & 

\begin{minipage}{.9\textwidth}
\begin{minted}[breaklines]{python}
def __init__(self,* args,** kwargs):
        self.args = args
        self.kwargs = kwargs
\end{minted}
\end{minipage}
\\
\\
\hline
\\0 & 

\begin{minipage}{.9\textwidth}
\begin{minted}[breaklines]{python}
def __init__(self,* args,** kwargs):

\end{minted}
\end{minipage}
\\
\\
\hline
\\
 \multicolumn{2}{c}{\textbf{Sequences sampled from a model fine-tuned using Ziegler}} \\  
1 & 

\begin{minipage}{.9\textwidth}
\begin{minted}[breaklines]{python}
def __init__(self,* args,** kwargs):
    self.args = args
    self.kwargs = kwargs
    self.kwargs_headers_headers_headers_headers_headers_headers_headers_headers_headers_headers_headers_headers_headers
\end{minted}
\end{minipage}
\\
\\
\hline
\\1 & 

\begin{minipage}{.9\textwidth}
\begin{minted}[breaklines]{python}
def __init__(self,* args,** kwargs):
        self.args = args
        self.kwargs = kwargs
        self.kwargs = kwargs
        self.kwargs = kwargs
        self.kwargs = kwargs
        self.kwargs = kwargs
        self.kwargs = kwargs
        self.kwargs = kwargs
\end{minted}
\end{minipage}
\\
\\
\hline
\\1 & 

\begin{minipage}{.9\textwidth}
\begin{minted}[breaklines]{python}
def __init__(self,* args,** kwargs):
        self.args = args
        self.args_args_args_args_args_args_args_args_args_args_args_args_args_args_args_args_args_args_args_args_args_
\end{minted}
\end{minipage}
\\
\\
\hline
\\
 \multicolumn{2}{c}{\textbf{Sequences sampled from a model fine-tuned using Reinforce}} \\  
1 & 

\begin{minipage}{.9\textwidth}
\begin{minted}[breaklines]{python}
def __init__(self,* args,** kwargs):
    self.typer = typer
\end{minted}
\end{minipage}
\\
\\
\hline
\\1 & 

\begin{minipage}{.9\textwidth}
\begin{minted}[breaklines]{python}
def __init__(self,* args,** kwargs):
    self.args = args
    self._unique_args = self._unique_args
\end{minted}
\end{minipage}
\\
\\
\hline
\\1 & 

\begin{minipage}{.9\textwidth}
\begin{minted}[breaklines]{python}
def __init__(self,* args,** kwargs):
    self.args = args
\end{minted}
\end{minipage}
\\
\\
\bottomrule
\end{tabular}
\caption{\small{Samples obtained from $\pit(\cdot|c)$ with $c =$ \texttt{def \_\_init\_\_(self,* args,** kwargs)} fine-tuned to satisfy a compilability constraint}\label{tab:code_1_samples}}
\end{table*}

\begin{table*}[h]
\tiny
\begin{tabular}{l|l}
\textbf{$b(x,c)$} & \textbf{$[x,c]$} \\ 
\midrule

 \multicolumn{2}{c}{\textbf{Sequences sampled from the original model $a$}} \\  
1 & 

\begin{minipage}{.9\textwidth}
\begin{minted}[breaklines]{python}
def __iter__(self):
          from models import model
          params = model.take_model_parameters()
          is_float = model.is_float_pos(not params)
          solvent = model.penesannex(params, is_float)
          solver_params = model.torneyred(params)
          solver_nwm(params)
\end{minted}
\end{minipage}
\\
\\
\hline
\\1 & 

\begin{minipage}{.9\textwidth}
\begin{minted}[breaklines]{python}
def __iter__(self):
    """
    Compare transformations stored in memory
    """
    if self.getFloat():
        self.reverseTransform()
        return list(self)
\end{minted}
\end{minipage}
\\
\\
\hline
\\1 & 

\begin{minipage}{.9\textwidth}
\begin{minted}[breaklines]{python}
def __iter__(self):
        self.filemanager()
\end{minted}
\end{minipage}
\\
\\
\hline
\\
 \multicolumn{2}{c}{\textbf{Sequences sampled from a model fine-tuned using Conditional DPG}} \\  
0 & 

\begin{minipage}{.9\textwidth}
\begin{minted}[breaklines]{python}
def __iter__(self):
            self.message = _L"{} a multidimensional array java geogrifier\n${}
            { }
        """ 
        if len(self.string_items) >= strlen(self.values):
            self.message += "[0] an empty list or json object\n\n"
            self.key
\end{minted}
\end{minipage}
\\
\\
\hline
\\1 & 

\begin{minipage}{.9\textwidth}
\begin{minted}[breaklines]{python}
def __iter__(self):
    self.container = None
    self.desc = None
    self.interview_name = None
    self.size_displacement_matrix = None
\end{minted}
\end{minipage}
\\
\\
\hline
\\0 & 

\begin{minipage}{.9\textwidth}
\begin{minted}[breaklines]{python}
def __iter__(self):
    """Destructor for test pairs.
\end{minted}
\end{minipage}
\\
\\
\hline
\\
 \multicolumn{2}{c}{\textbf{Sequences sampled from a model fine-tuned using DPG}} \\  
1 & 

\begin{minipage}{.9\textwidth}
\begin{minted}[breaklines]{python}
def __iter__(self):
    return self.__iter__(self)
\end{minted}
\end{minipage}
\\
\\
\hline
\\1 & 

\begin{minipage}{.9\textwidth}
\begin{minted}[breaklines]{python}
def __iter__(self):
        return [self.genState[0] for _ in self.stateNodes]
\end{minted}
\end{minipage}
\\
\\
\hline
\\1 & 

\begin{minipage}{.9\textwidth}
\begin{minted}[breaklines]{python}
def __iter__(self):
        created = True
        for i in xrange(0, len(self.callover)):
            self.callover[scoped_async_callover(i)] = created
\end{minted}
\end{minipage}
\\
\\
\hline
\\
 \multicolumn{2}{c}{\textbf{Sequences sampled from a model fine-tuned using Ziegler}} \\  
1 & 

\begin{minipage}{.9\textwidth}
\begin{minted}[breaklines]{python}
def __iter__(self):
        self.items_items_items_items_items_items_items_items_items_items_items_items_items_items_items_items_items_items
\end{minted}
\end{minipage}
\\
\\
\hline
\\1 & 

\begin{minipage}{.9\textwidth}
\begin{minted}[breaklines]{python}
def __iter__(self):
    self.iterator = iterator_ops_ops_ops_ops_ops_ops_ops_ops_ops_ops_ops_ops_ops_ops_ops_ops_ops_ops_ops_ops_ops_ops_ops_ops_ops_ops_ops_ops
\end{minted}
\end{minipage}
\\
\\
\hline
\\1 & 

\begin{minipage}{.9\textwidth}
\begin{minted}[breaklines]{python}
def __iter__(self):
    self._iter_names_children_children_children_children_children_children_children_children_children_children_children
\end{minted}
\end{minipage}
\\
\\
\hline
\\
 \multicolumn{2}{c}{\textbf{Sequences sampled from a model fine-tuned using Reinforce}} \\  
1 & 

\begin{minipage}{.9\textwidth}
\begin{minted}[breaklines]{python}
def __iter__(self):
        self.next()
\end{minted}
\end{minipage}
\\
\\
\hline
\\1 & 

\begin{minipage}{.9\textwidth}
\begin{minted}[breaklines]{python}
def __iter__(self):
    self._iterator = self
\end{minted}
\end{minipage}
\\
\\
\hline
\\1 & 

\begin{minipage}{.9\textwidth}
\begin{minted}[breaklines]{python}
def __iter__(self):
    self.__iter__()
\end{minted}
\end{minipage}
\\
\\
\bottomrule
\end{tabular}
\caption{\small{Samples obtained from $\pit(\cdot|c)$ with $c =$ \texttt{def \_\_iter\_\_(self)} fine-tuned to satisfy a compilability constraint}\label{tab:code_2_samples}}
\end{table*}

\begin{table*}[h]
\tiny
\begin{tabular}{l|l}
\textbf{$b(x,c)$} & \textbf{$[x,c]$} \\ 
\midrule

 \multicolumn{2}{c}{\textbf{Sequences sampled from the original model $a$}} \\  
0 & 

\begin{minipage}{.9\textwidth}
\begin{minted}[breaklines]{python}
def __init__(self):
    """ mutation() method.
\end{minted}
\end{minipage}
\\
\\
\hline
\\0 & 

\begin{minipage}{.9\textwidth}
\begin{minted}[breaklines]{python}
def __init__(self):
    self.vcam = cham OLEDeaderCenterPythRHAPUV()
    self.vcam_rc = bool_ ("with_ DisneySaudiDisney) Official Created Source Title and Updated Collection
    self.volume = float(1000)
    self.title = float(params()[3])
\end{minted}
\end{minipage}
\\
\\
\hline
\\0 & 

\begin{minipage}{.9\textwidth}
\begin{minted}[breaklines]{python}
def __init__(self):
        """
        Returns the inner type that is of the proper type when performing
        self-self-transpose().
        """
        self.th = "arg1",
\end{minted}
\end{minipage}
\\
\\
\hline
\\
 \multicolumn{2}{c}{\textbf{Sequences sampled from a model fine-tuned using Conditional DPG}} \\  
0 & 

\begin{minipage}{.9\textwidth}
\begin{minted}[breaklines]{python}
def __init__(self):
    self.tile_flag_handler_out_A = True
    self.tile_flag_handler_out_B = True
    self.layer_list_out_A = self.tile_flag_invalid=2
    self.layer_list_out_B = self.tile_flag_handler_out_A
    self.layer_select_with_dropdown_flat_flag_in=self.tile_flag_invalid=
\end{minted}
\end{minipage}
\\
\\
\hline
\\0 & 

\begin{minipage}{.9\textwidth}
\begin{minted}[breaklines]{python}
def __init__(self):
    return self.InitExceptaughtClient();
\end{minted}
\end{minipage}
\\
\\
\hline
\\1 & 

\begin{minipage}{.9\textwidth}
\begin{minted}[breaklines]{python}
def __init__(self):
    flood_handler = ScreenGroup.set_buffer_manager_function(buffers, None)
    heavy = _charging_class(self, fill_handler)
\end{minted}
\end{minipage}
\\
\\
\hline
\\
 \multicolumn{2}{c}{\textbf{Sequences sampled from a model fine-tuned using DPG}} \\  
1 & 

\begin{minipage}{.9\textwidth}
\begin{minted}[breaklines]{python}
def __init__(self):
    class App(QtWndrixApplication):
        Slipped: bool = QtWnd technique  # Member of gtk.SafePlacement floated
        delegate: NamedConstraint: QListPlacementElementDelegateDelegate
\end{minted}
\end{minipage}
\\
\\
\hline
\\0 & 

\begin{minipage}{.9\textwidth}
\begin{minted}[breaklines]{python}
def __init__(self):
    def gqlary(self):
        presentation_data = (["$parent", " Imperium", "Segaron", "la datos de 214000 "],
                                               [["Lamos en su inicio", 0, "Por supuesto", 0],
               
\end{minted}
\end{minipage}
\\
\\
\hline
\\0 & 

\begin{minipage}{.9\textwidth}
\begin{minted}[breaklines]{python}
def __init__(self):
	self.msg = {}
	#Placeholders
\end{minted}
\end{minipage}
\\
\\
\hline
\\
 \multicolumn{2}{c}{\textbf{Sequences sampled from a model fine-tuned using Ziegler}} \\  
0 & 

\begin{minipage}{.9\textwidth}
\begin{minted}[breaklines]{python}
def __init__(self):
                                   name=SHA_['base'].__name__().strip()
                                   message=Message(path=_md5. 'value'))
\end{minted}
\end{minipage}
\\
\\
\hline
\\1 & 

\begin{minipage}{.9\textwidth}
\begin{minted}[breaklines]{python}
def __init__(self):
    as_printer = asv(0)
    i = 3
    symbol = _sparse_symbol(0)
\end{minted}
\end{minipage}
\\
\\
\hline
\\0 & 

\begin{minipage}{.9\textwidth}
\begin{minted}[breaklines]{python}
def __init__(self):
        self.identity_dont_print_view = 'Identities'
\end{minted}
\end{minipage}
\\
\\
\hline
\\
 \multicolumn{2}{c}{\textbf{Sequences sampled from a model fine-tuned using Reinforce}} \\  
0 & 

\begin{minipage}{.9\textwidth}
\begin{minted}[breaklines]{python}
def __init__(self):
        self.dataset = TensorsRotatedDeviceDataset()
        self.start = None
        self.dataset.dataset = TensorsRotatedPose(self, 100)
\end{minted}
\end{minipage}
\\
\\
\hline
\\1 & 

\begin{minipage}{.9\textwidth}
\begin{minted}[breaklines]{python}
def __init__(self):
    tf.__init__(self)
\end{minted}
\end{minipage}
\\
\\

\bottomrule
\end{tabular}
\caption{\small{Samples obtained from $\pit(\cdot|c)$ with $c =$ \texttt{def \_\_init\_\_(self)} fine-tuned to satisfy a PEP8 constraint}\label{tab:code_pep8_0_samples}}
\end{table*}

\begin{table*}[h]
\tiny
\begin{tabular}{l|l}
\textbf{$b(x,c)$} & \textbf{$[x,c]$} \\ 
\midrule

 \multicolumn{2}{c}{\textbf{Sequences sampled from the original model $a$}} \\  
0 & 

\begin{minipage}{.9\textwidth}
\begin{minted}[breaklines]{python}
def __init__(self,* args,** kwargs):
        it = lambda: args
        it(None)
\end{minted}
\end{minipage}
\\
\\
\hline
\\0 & 

\begin{minipage}{.9\textwidth}
\begin{minted}[breaklines]{python}
def __init__(self,* args,** kwargs):

\end{minted}
\end{minipage}
\\
\\
\hline
\\0 & 

\begin{minipage}{.9\textwidth}
\begin{minted}[breaklines]{python}
def __init__(self,* args,** kwargs):
    raise RuntimeError('Cannot create new thread');
\end{minted}
\end{minipage}
\\
\\
\hline
\\
 \multicolumn{2}{c}{\textbf{Sequences sampled from a model fine-tuned using Conditional DPG}} \\  
0 & 

\begin{minipage}{.9\textwidth}
\begin{minted}[breaklines]{python}
def __init__(self,* args,** kwargs):
    dgram = dgram_preprocessor(self.parser,* args,**kwargs)
    self.node = __defaultdict__[node]
    kwargs = kwargs
    self.classpath = "%(classpath)s//%(pathname)s//"
    self.param = classpath()
    self.property = classpath()
    self.sys.exists = False
    return self
\end{minted}
\end{minipage}
\\
\\
\hline
\\0 & 

\begin{minipage}{.9\textwidth}
\begin{minted}[breaklines]{python}
def __init__(self,* args,** kwargs):
        self.app = comprom::Configuration(conf())
        conf = self.confswite(args.get('app'), args.get('name', ''), conf)
\end{minted}
\end{minipage}
\\
\\
\hline
\\0 & 

\begin{minipage}{.9\textwidth}
\begin{minted}[breaklines]{python}
def __init__(self,* args,** kwargs):

\end{minted}
\end{minipage}
\\
\\
\hline
\\
 \multicolumn{2}{c}{\textbf{Sequences sampled from a model fine-tuned using DPG}} \\  
0 & 

\begin{minipage}{.9\textwidth}
\begin{minted}[breaklines]{python}
def __init__(self,* args,** kwargs):

\end{minted}
\end{minipage}
\\
\\
\hline
\\0 & 

\begin{minipage}{.9\textwidth}
\begin{minted}[breaklines]{python}
def __init__(self,* args,** kwargs):

\end{minted}
\end{minipage}
\\
\\
\hline
\\0 & 

\begin{minipage}{.9\textwidth}
\begin{minted}[breaklines]{python}
def __init__(self,* args,** kwargs):
    self.tip   = {}
    self.buffer = ''
    self.clean_of_this  = ''
    self.tip.set_value(op_type())
    self.tip.set_value(ostream(f" excerpting this", oproc = self.poetrip_extrapol))
\end{minted}
\end{minipage}
\\
\\
\hline
\\
 \multicolumn{2}{c}{\textbf{Sequences sampled from a model fine-tuned using Ziegler}} \\  
0 & 

\begin{minipage}{.9\textwidth}
\begin{minted}[breaklines]{python}
def __init__(self,* args,** kwargs):
        U particles = sorted(p.get_particles(c4))
        kwargs.append(p)
        if self.like_class:
            p.like_class = kwargs[0]
\end{minted}
\end{minipage}
\\
\\
\hline
\\0 & 

\begin{minipage}{.9\textwidth}
\begin{minted}[breaklines]{python}
def __init__(self,* args,** kwargs):
        if not c.c.service.defaultService():
            args.favoriter(c.service.defaultService(c.pass.name))
        if not c.simpleAddr():
            d.favoriter(c.simpleAddr)
        if c.simpleAddr():
            c.gop.services.import_service(d
\end{minted}
\end{minipage}
\\
\\
\hline
\\0 & 

\begin{minipage}{.9\textwidth}
\begin{minted}[breaklines]{python}
def __init__(self,* args,** kwargs):
            self.init_args(args,kwargs)
            self.base_args = base_args
\end{minted}
\end{minipage}
\\
\\
\hline
\\
 \multicolumn{2}{c}{\textbf{Sequences sampled from a model fine-tuned using Reinforce}} \\  
0 & 

\begin{minipage}{.9\textwidth}
\begin{minted}[breaklines]{python}
def __init__(self,* args,** kwargs):
    """ In constructor, bubbles can be used accordingly.
\end{minted}
\end{minipage}
\\
\\
\hline
\\0 & 

\begin{minipage}{.9\textwidth}
\begin{minted}[breaklines]{python}
def __init__(self,* args,** kwargs):
    self.akpo_be(akpo_key,**args)
    self.appo_a(a,r)\
     .name(dfn)//type(dfn)-1
    sdt=self.akpo_ fishing(a//a[0],r@a[0])[0]
    sd=self.akpo_ a
    dfn=self.akpo_df(ado(dfn)),
    a=self.akpo_a(rs):
\end{minted}
\end{minipage}
\\
\\
\hline
\\0 & 

\begin{minipage}{.9\textwidth}
\begin{minted}[breaklines]{python}
def __init__(self,* args,** kwargs):
    from __future__ import print_function
\end{minted}
\end{minipage}
\\
\\
\bottomrule
\end{tabular}
\caption{\small{Samples obtained from $\pit(\cdot|c)$ with $c =$ \texttt{def \_\_init\_\_(self,* args,** kwargs)} fine-tuned to satisfy a PEP8 constraint}\label{tab:code_pep8_1_samples}}
\end{table*}

\begin{table*}[h]
\tiny
\begin{tabular}{l|l}
\textbf{$b(x,c)$} & \textbf{$[x,c]$} \\ 
\midrule

 \multicolumn{2}{c}{\textbf{Sequences sampled from the original model $a$}} \\  
1 & 

\begin{minipage}{.9\textwidth}
\begin{minted}[breaklines]{python}
def __repr__(self):
    return "__repr__"
\end{minted}
\end{minipage}
\\
\\
\hline
\\0 & 

\begin{minipage}{.9\textwidth}
\begin{minted}[breaklines]{python}
def __repr__(self):
        raise StopIteration("operator"
                               "(mut_index_t) not an iterable", self)
        self.ignoreSelfToken(self)
\end{minted}
\end{minipage}
\\
\\
\hline
\\0 & 

\begin{minipage}{.9\textwidth}
\begin{minted}[breaklines]{python}
def __repr__(self):
    self.memory = self._object_create_memory
    self.mainfile = None
    def ref() -> keyboard:
        raise self.error('"—<–' pass)
\end{minted}
\end{minipage}
\\
\\
\hline
\\
 \multicolumn{2}{c}{\textbf{Sequences sampled from a model fine-tuned using Conditional DPG}} \\  
0 & 

\begin{minipage}{.9\textwidth}
\begin{minted}[breaklines]{python}
def __repr__(self):
    """standard
    :param self:
\end{minted}
\end{minipage}
\\
\\
\hline
\\0 & 

\begin{minipage}{.9\textwidth}
\begin{minted}[breaklines]{python}
def __repr__(self):
    if __name__ == '__main__':
        return self.os_main()
    else:
        return self.req.as_raw_object(
            __name__,
            :method => 'GET',
            accept_parser=self.parser, accept_kwargs=self.accept_kwargs,
        
\end{minted}
\end{minipage}
\\
\\
\hline
\\1 & 

\begin{minipage}{.9\textwidth}
\begin{minted}[breaklines]{python}
def __repr__(self):
    return self._name + '_' + self._rubyblob.result
\end{minted}
\end{minipage}
\\
\\
\hline
\\
 \multicolumn{2}{c}{\textbf{Sequences sampled from a model fine-tuned using DPG}} \\  
1 & 

\begin{minipage}{.9\textwidth}
\begin{minted}[breaklines]{python}
def __repr__(self):
    self.loadstrli()
\end{minted}
\end{minipage}
\\
\\
\hline
\\0 & 

\begin{minipage}{.9\textwidth}
\begin{minted}[breaklines]{python}
def __repr__(self):
    """
    Return the normal to ancient_api._has_proto(self) property of the
    class. To unlock up the hyperlabels, use __set__() instead.
    """
    old_l = lazy_data_type(l)
    found = set()
    raw = lazy_data_type( rawToUpperNone, "need full valid types for serialized
    datatypes" ) where idxs = raw; '1' if raw and hasattr( raw
\end{minted}
\end{minipage}
\\
\\
\hline
\\0 & 

\begin{minipage}{.9\textwidth}
\begin{minted}[breaklines]{python}
def __repr__(self):
    """Revert to the previously returned Signing direction.
\end{minted}
\end{minipage}
\\
\\
\hline
\\
 \multicolumn{2}{c}{\textbf{Sequences sampled from a model fine-tuned using Ziegler}} \\  
0 & 

\begin{minipage}{.9\textwidth}
\begin{minted}[breaklines]{python}
def __repr__(self):
    with self:
        self.__repr__[ 0.0] = 1
\end{minted}
\end{minipage}
\\
\\
\hline
\\1 & 

\begin{minipage}{.9\textwidth}
\begin{minted}[breaklines]{python}
def __repr__(self):
    _ = self.name
\end{minted}
\end{minipage}
\\
\\
\hline
\\0 & 

\begin{minipage}{.9\textwidth}
\begin{minted}[breaklines]{python}
def __repr__(self):
                                      y
                                                        
\end{minted}
\end{minipage}
\\
\\
\hline
\\
 \multicolumn{2}{c}{\textbf{Sequences sampled from a model fine-tuned using Reinforce}} \\  
1 & 

\begin{minipage}{.9\textwidth}
\begin{minted}[breaklines]{python}
def __repr__(self):
    self = self._self
    self._self = self._self
    return self
\end{minted}
\end{minipage}
\\
\\
\hline
\\0 & 

\begin{minipage}{.9\textwidth}
\begin{minted}[breaklines]{python}
def __repr__(self):
    __repr__(self):
    v = self.__repr__(self)
    if v not in __repr__:
        __repr__(self, v)
    return v
\end{minted}
\end{minipage}
\\
\\
\hline
\\0 & 

\begin{minipage}{.9\textwidth}
\begin{minted}[breaklines]{python}
def __repr__(self):
    print("\n...\n....").stack(self.dim[0], self.dimensional[0][0], sizeof(self) * self.dimensional[0])
\end{minted}
\end{minipage}
\\
\\
\bottomrule
\end{tabular}
\caption{\small{Samples obtained from $\pit(\cdot|c)$ with $c =$ \texttt{def \_\_repr\_\_(self)} fine-tuned to satisfy a PEP8 constraint}\label{tab:code_pep8_2_samples}}
\end{table*}

\end{document}